\documentclass[11pt]{article}

\usepackage[final]{acl}

\usepackage{times}
\usepackage{latexsym}

\usepackage[T1]{fontenc}

\usepackage[utf8]{inputenc}

\usepackage{microtype}

\usepackage{inconsolata}
\usepackage{amssymb}
\usepackage{booktabs}
\usepackage{array}

\usepackage{graphicx}

\title{FreqBLiMP: Frequency-Controlled Minimal Pairs Reveal \\ Robustness and Fragility of LLMs Under Lexical Rarity} 

\author{Tyrone White \and Yuki Arase \\
  School of Computing, Institute of Science Tokyo \\
  Tokyo, Japan \\
  \texttt{white.q.a4e7@m.isct.ac.jp} \quad
  \texttt{arase@c.titech.ac.jp}}

\begin{document}
\maketitle
\begin{abstract}
Minimal-pair benchmarks such as BLiMP evaluate linguistic knowledge by
testing whether language models (LMs) prefer acceptable sentences over minimally
different unacceptable ones. 
However, these benchmarks largely ignore lexical frequency variation, despite lexical frequency being a pervasive and highly skewed property of natural language use. Consequently, existing evaluations do not test whether grammatical preferences remain stable when contrasts involve rare lexical items.
We introduce \textsc{FreqBLiMP}, a
frequency-controlled extension of BLiMP that regenerates all $67$ paradigms under
explicit Zipf-frequency regimes while preserving each minimal-pair's grammatical contrast. 
Evaluating multiple open-weight LLM families across scales, we find that decreasing lexical frequency produces a consistent,
monotonic decrease in sentence likelihood, but only a modest reduction in overall contrastive acceptability accuracy. 
However, this aggregate stability masks substantial variation across linguistic phenomena, with LLMs remaining robust on overt morphosyntactic generalization while degrading on phenomena that require lemma-specific information.
\end{abstract}

\section{Introduction}
\label{sec:intro}

Large language models (LLMs) perform well on many tests of linguistic
knowledge, but it remains unclear whether this performance reflects abstraction over grammatical structure rather than familiarity with the lexical material  \citep{gulordava2018colorless,newman2021refining,lasri2022does}. Targeted minimal-pair benchmarks probe this question by
presenting models with closely matched grammatically acceptable and
unacceptable sentences and testing whether the model assigns higher probability
to the acceptable one
\citep{linzen2016assessing,marvin2018targeted,warstadt2020blimp}. BLiMP
\citep{warstadt2020blimp} systematizes this paradigm across $67$ English
paradigms spanning morphology, syntax, and syntax--semantics, and has become a standard benchmark for evaluating linguistic acceptability in LMs.

However, existing minimal-pair benchmarks largely ignore lexical frequency variation. Because benchmark items are predominantly constructed from common lexical items, they do not test whether grammatical preferences remain stable when the same contrasts involve rare words. 
This limitation is important because natural language exhibits a highly skewed frequency distribution \citep{zipf1949human,piantadosi2014zipf}, and rare words
differ systematically from frequent ones in ways that may affect model behavior, including reduced contextual coverage, increased subword fragmentation, and less prototypical lexical-semantic environments. Consequently, strong performance on standard minimal-pair benchmarks may not indicate robust grammatical generalization across the lexical frequency spectrum. 
A model that correctly prefers (1a) over (1b) may fail to preserve the same distinction when the contrast is instantiated with rarer lexical material, as in (2) and (3).

\begin{quote}
\noindent (1a) Victoria finds this book.\\
(1b) *Victoria finds these book.\\[0.25em]
(2a/b) This taxonomy alters these/*this dealerships.\\
(3a/b) The jockeys antagonize this/*these reformist.
\end{quote}

\noindent
At the same time, lower probability for a rare-word sentence does not by itself show
a loss of grammatical sensitivity: a model may penalize both members of a minimal pair globally while still correctly preferring the acceptable sentence. Evaluating robustness to lexical rarity therefore requires varying lexical frequency while preserving the underlying acceptability contrast.

To address this problem, we introduce \textsc{FreqBLiMP}, a frequency-controlled extension of BLiMP
that regenerates its template-generated minimal pairs across explicit Zipf-frequency regimes while preserving their grammatical acceptability contrasts.
We expand the generator vocabulary with additional
nouns, verbs, and adjectives, annotate them with the syntactic and semantic
features required by BLiMP's templates, and sample eligible lexical items
based on frequency estimates.
This yields parallel high-frequency, low-frequency, and very-low-frequency
benchmark regimes while preserving the minimal-pair structure and grammatical contrasts of each paradigm.

We evaluate open-weight LLMs using standard sentence likelihood and two
acceptability-framed scoring methods. Decreasing lexical frequency
consistently lowers sentence likelihood as expected but produces only a modest reduction
in aggregate contrastive accuracy. This aggregate stability, however, masks substantial
heterogeneity across paradigms. The clearest declines occur when the contrast
requires information associated with particular lexical items, including
verbal argument-frame properties \citep{levin1993english}, animacy, and
lexical gender. By contrast, contrasts supported by overt morphosyntactic
cues, especially agreement morphology, remain stable or sometimes improve
under rare lexicalization, revealing a linguistic dissociation in how grammatical constraints generalize across the lexical frequency spectrum.


Our contributions are threefold:
(i) \textsc{FreqBLiMP}, a framework for adding lexical-frequency control to
BLiMP-style minimal-pair evaluation;
(ii) benchmark regimes that systematically vary lexical rarity while
preserving the targeted grammatical contrasts; and
(iii) empirical evidence for a linguistic dissociation under lexical rarity:
contrasts requiring lemma-specific lexical information degrade more strongly,
whereas contrasts supported by overt morphosyntactic cues remain comparatively
robust.
\textsc{FreqBLiMP}, its generation pipeline, and evaluation code are publicly
available at \url{https://github.com/TimeTravelerTy/freqblimp-generation}.

\section{Related Work}
\label{sec:related}

\paragraph{Targeted grammatical evaluation.}
A common way to test grammatical knowledge in LMs is to evaluate
their preferences over minimal pairs: closely matched sentences that differ in
acceptability and isolate a targeted contrast. This paradigm has been used to
study agreement, reflexives, negative-polarity licensing, filler--gap
dependencies, and other structure-sensitive phenomena
\citep[e.g.,][]{linzen2016assessing,marvin2018targeted,wilcox2018rnn,
gulordava2018colorless,warstadt2020blimp,gauthier2020syntaxgym}.
BLiMP \citep{warstadt2020blimp} scales this approach to $67$
template-generated English paradigms spanning morphology, syntax, and
syntax--semantics, evaluating whether models assign higher probability to the
acceptable sentence in each pair. Unlike CoLA-style single-sentence evaluation
\citep{warstadt2019neural}, minimal-pair evaluation does not require training a
supervised acceptability classifier and provides tighter control over the
targeted grammatical contrast.

\paragraph{Frequency and the long tail.}
Word frequency is a pervasive property of natural language: a small number of
types dominate token counts, while most lexical items lie in the long tail
\citep{zipf1949human,piantadosi2014zipf}. Prior work has shown that lexical
frequency and long-tail structure affect factual knowledge, surprisal estimates,
generation diversity, and knowledge of low-frequency or idiosyncratic
constructions
\citep{kandpal2023large,oh2024frequency,dohmatob2024model,
shumailov2024ai,ju2025domain,mahowald2023discerning}. Most closely related to
our setting, \citet{algayres2025longtail} introduce LT-Swap, a corpus-specific
benchmark for evaluating semantic and syntactic use of rare words through
minimal-pair swaps. Their goal is to measure rare-word acquisition relative to a
known pretraining distribution. 
In contrast, our goal is to test whether BLiMP-style
grammatical acceptability preferences remain stable when the same
template-generated contrasts are instantiated under controlled Zipf-frequency
regimes.

\paragraph{Positioning of our work.}
Prior work on grammatical acceptability and lexical frequency has largely proceeded independently. Existing minimal-pair benchmarks focus on evaluating grammatical sensitivity under controlled contrasts, while work on lexical frequency and long-tail behavior primarily examines the usage and acquisition of rare words. Related work on gradient acceptability has also examined how lexical frequency and sentence length affect the relationship between LM probability and human judgments \citep{lau2017grammaticality,tjuatja-etal-2025-goes}. Our work connects these lines of research by investigating how lexical frequency affects LM acceptability judgments under controlled minimal-pair evaluation.

\section{FreqBLiMP}
\label{sec:construction}

\textsc{FreqBLiMP} regenerates all of the $67$ BLiMP paradigms
\citep{warstadt2020blimp} under explicit lexical-frequency constraints. For
each paradigm, we generate $1{,}000$ unique minimal pairs in each of three
frequency regimes, yielding $67{,}000$ pairs per regime. The paradigm metadata,
evaluation format, and acceptability contrast follow BLiMP; only eligible
content words are resampled, with invalid or duplicate generations discarded.



\subsection{Frequency-Controlled Sampling}
\label{subsec:frequency-control}

We apply frequency control to content-word positions (nouns, verbs, and
adjectives) while leaving function words and proper nouns unchanged. Lexical
frequency is measured using Zipf estimates from \texttt{wordfreq} \citep{speer2022wordfreq}, where one Zipf point corresponds to a tenfold
frequency difference, and Zipf~3 is approximately one occurrence per million
words. Candidates are filtered by the regime-specific windows in
Table~\ref{tab:freq-regimes} and sampled uniformly from the filtered pool. The
windows were chosen empirically to produce clearly separated levels of lexical
familiarity while still leaving enough valid candidates for constrained BLiMP
paradigms. Frequency checks are applied to the surface form used by the
template, so inflected forms such as \textit{walks}, \textit{walked}, or
\textit{walking} are filtered by their own Zipf frequency rather than that of
the lemma.

\begin{table}[t]
\centering
\begin{tabular}{lc}
\hline
\textbf{Regime} & \textbf{Zipf range} \\
\hline
Head & 3.5--5.5 \\
Tail & 2.4--3.2 \\
XTail & 1.2--2.2 \\
\hline
\end{tabular}
\caption{Zipf-frequency ranges used in \textsc{FreqBLiMP} generation.}
\label{tab:freq-regimes}
\end{table}

\subsection{Lexical Overlay}
\label{subsec:lexical-overlay}

The original BLiMP lexicon is too small to support systematic rare-word
generation, especially for semantically constrained nouns and verbs. We
therefore construct a lexical overlay that expands BLiMP's noun, adjective,
and verb inventories while preserving the feature structure expected by the
original generators. 

New lemmas are drawn primarily from Open English WordNet \citep{mccrae-etal-2019-english}; for human-denoting nouns, we additionally use
Wikidata-derived person and gender information, filtering out potentially offensive terms \citep{vrandecic-krotzsch-2014-wikidata}. For verbs, we use frame information from VerbNet
\citep{kipper_schuler_2005_verbnet}, WordNet, corpus checks, and manual review. Candidates are filtered for basic suitability, mapped to
BLiMP-compatible syntactic or semantic bundles, and inserted into existing
lexical templates. These bundles encode properties such as syntactic category,
inflectional features, agreement behavior, animacy, countability, semantic
class, selectional restrictions, and verbal argument frame.

Some BLiMP paradigms require more specific lexical behavior than can be reliably obtained from broad resource-based sampling. For these cases, including control, raising, tough-adjective constructions, passivization, causative--inchoative alternations, and object drop, we use manually curated predicate lists, restricting candidates to predicates verified to license the relevant alternation, so that lexical expansion preserves the intended contrast. When a curated list contains too few in-window predicates, we use nearest-frequency fallbacks. This lowers coverage for several control/raising and tough-construction paradigms, which we therefore treat as descriptive. During development, we manually inspected generated samples to identify recurring lexicalization errors and refine filters, templates, and these curated lists. Appendix~\ref{app:lexical-overlay-implementation} specifies implementation details and reports in-window coverage.

Table~\ref{tab:freqblimp-examples} gives representative generated pairs,
illustrating how \textsc{FreqBLiMP} keeps the BLiMP minimal-pair contrast fixed
while varying the lexical material used to realize each construction.

\begin{table*}[t]
\centering
\footnotesize
\setlength{\tabcolsep}{4pt}
\begin{tabular}{@{}p{0.18\textwidth}p{0.10\textwidth}p{0.33\textwidth}p{0.33\textwidth}@{}}
\toprule
\textbf{Paradigm} & \textbf{Regime} & \textbf{Acceptable sentence} & \textbf{Unacceptable sentence} \\
\midrule
passive 1 & Head &
\textit{Andrea isn't \underline{avoided} by Irene.} &
$^\ast$\textit{Andrea isn't \underline{responded} by Irene.} \\
passive 1 & Tail &
\textit{All fitters are \underline{lauded} by most laymen.} &
$^\ast$\textit{All fitters are \underline{toiled} by most laymen.} \\
passive 1 & XTail &
\textit{These enterprisers were \underline{lugged} by the gasman.} &
$^\ast$\textit{These enterprisers were \underline{lolled} by the gasman.} \\
principle A domain 3 & Head &
\textit{\underline{This ex-husband} feels \underline{the daughter} would conduct herself.} &
$^\ast$\textit{\underline{The daughter} feels \underline{this ex-husband} would conduct herself.} \\
principle A domain 3 & Tail &
\textit{\underline{A supergirl} imagines \underline{some widower} deceives himself.} &
$^\ast$\textit{\underline{Some widower} imagines \underline{a supergirl} deceives himself.} \\
principle A domain 3 & XTail &
\textit{\underline{Every songster} imagines \underline{a bhikkhuni} edifies herself.} &
$^\ast$\textit{\underline{A bhikkhuni} imagines \underline{every songster} edifies herself.} \\
\bottomrule
\end{tabular}
\caption{
Representative \textsc{FreqBLiMP} minimal pairs from argument structure and
binding paradigms. Each row preserves the target BLiMP contrast while
resampling eligible content words under the indicated Zipf regime. Underlining
marks the passive-licensing verb or the subject phrases whose order determines
the reflexive's local antecedent.
}
\label{tab:freqblimp-examples}
\end{table*}

\subsection{FreqBLiMP Profile}
\label{sec:diagnostics}

\begin{figure}[t]
  \centering
  \includegraphics[width=\columnwidth]{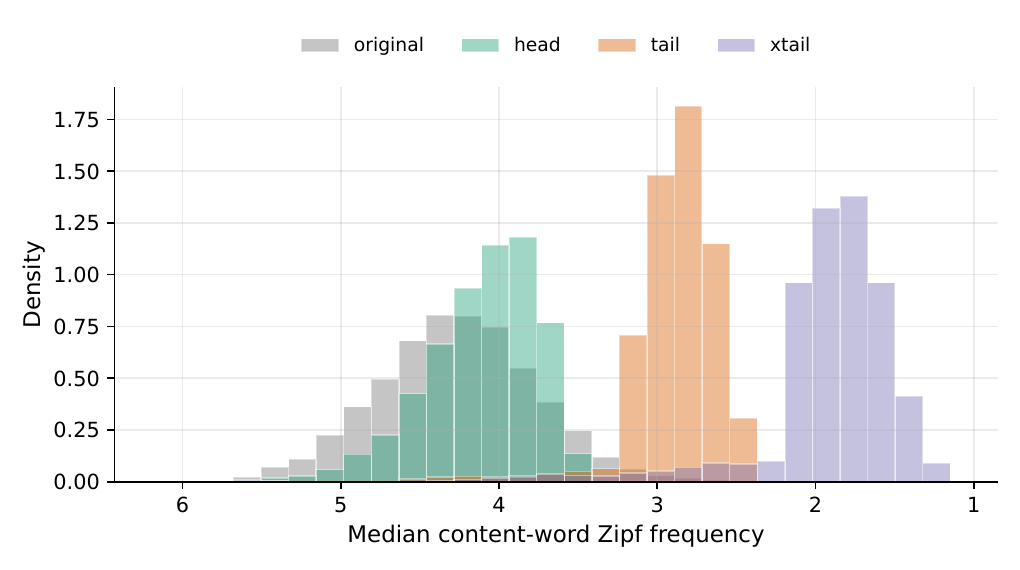}
  \caption{
    Realized content-word Zipf distributions for original BLiMP and generated
    \textsc{FreqBLiMP} regimes. The $x$-axis is reversed so that lexical
    frequency decreases from left to right.
  }
  \label{fig:zipf-dist}
\end{figure}

We next characterize the generated benchmark to verify that the frequency-controlled regimes are
lexically distinct, diverse, and comparable to original BLiMP in the intended
way. Figure~\ref{fig:zipf-dist} shows the realized content-word Zipf distributions
for original BLiMP and the three generated \textsc{FreqBLiMP} regimes. The
generated regimes are well separated: Head items occupy the high-frequency
range, Tail items center around Zipf~$\approx 2.9$, and XTail items
center below Zipf~$2$. 
The original BLiMP overlaps mostly with the
Head regime but is broader, reflecting the fact that its lexical material was not
frequency-controlled.

The generated regimes are also lexically diverse. Across frequency-controlled
content-word slots, they contain substantially more unique lemmas and surface
forms than original BLiMP, while the top-20 lemma share remains below $5\%$ in
all generated regimes (Table~\ref{tab:regime-lexical-profile}). We report the
full lexical-diversity statistics, including effective lemma counts, in
Appendix~\ref{app:lexical-diversity}.

\begin{table}[t]
\centering
\small
\begin{tabular}{lrrr}
\hline
\textbf{Dataset / regime} & \textbf{Median Zipf} & \textbf{Lemmas} & \textbf{Top-20} \\
\hline
Original BLiMP & 4.26 & 2,607 & 14.3 \\
Head & 4.05 & 6,304 & 2.9 \\
Tail & 2.85 & 10,583 & 4.3 \\
XTail & 1.85 & 15,111 & 4.0 \\
\hline
\end{tabular}
\caption{
Lexical profile by regime. Median Zipf is computed over realized
frequency-controlled content-word occurrences; Top-20 is the percentage of
those occurrences covered by the 20 most frequent lemmas.
}
\label{tab:regime-lexical-profile}
\end{table}


\section{Experimental Setup}
\label{sec:setup}

Our experiments test whether decreasing content-word frequency affects two
separable aspects of model behavior: overall sentence expectation and
contrastive grammatical preference.

\subsection{Models}
We evaluate a broad suite of open-weight LLMs spanning multiple
families and scales to compare the effects of model family, scale and post-training. 
Namely, the following models are covered.
\begin{itemize}
    \item Llama-3.1: 8B, 70B, 70B-Instruct 
    \item Qwen2.5: 7B, 72B, 72B-Instruct
    \item Gemma-4: E4B, 31B, 31B-it
    \item Mistral-7B-v0.1
\end{itemize}

\subsection{Scoring Methods}
Our primary evaluation follows the standard BLiMP minimal-pair likelihood
protocol. 
For each pair in \textsc{FreqBLiMP} denoted as $\mathcal{D}$, consisting of a grammatical sentence $g$ and an
ungrammatical sentence $b$, we compute left-to-right sentence log probability (referred to as LP readout),
\begin{equation}
\mathrm{LP}(s) = \sum_{t=1}^{T} \log p(x_t \mid x_{<t}),
\end{equation}
and count a model as correct when $\mathrm{LP}(g) > \mathrm{LP}(b)$. 
The accuracy is defined as 
\begin{equation}
\label{eq:accuracy}
\mathrm{Accuracy}
=
\frac{1}{|\mathcal{D}|}
\sum_{(g,b)\in\mathcal{D}}
\mathbb{I}\!\left[\mathrm{LP}(g) > \mathrm{LP}(b)\right].
\end{equation}
We also report the log-probability margin,
\begin{equation}
\label{eq:delta_lp}
\Delta \mathrm{LP} = \mathrm{LP}(g) - \mathrm{LP}(b),
\end{equation}
where larger positive values indicate a stronger preference for the grammatical
sentence.

Because direct sentence likelihood can conflate grammaticality with naturalness,
lexical expectation, and length, we additionally evaluate two prompted
probability-based alternatives motivated by \citet{ide-etal-2025-make}: Template LP and Yes/No scoring.
For Template LP, we embed each sentence in the same grammaticality frame:
``The following sentence is grammatically acceptable. [sentence]'' and compute LP readout. 
For Yes/No scoring, we place each sentence in a grammaticality question prompt,
asking whether the sentence is grammatically acceptable and instructing the
model to answer \textit{Yes} or \textit{No}. 
We read out the next-token probability mass assigned to single-token variants of \textit{Yes} and \textit{No}, and define $P_{\mathrm{yes}}(s)$ as the normalized probability mass assigned to \textit{Yes}.
A pair is counted as correct when
$P_{\mathrm{yes}}(g) > P_{\mathrm{yes}}(b)$.

We compute all scores for both base and instruction-tuned
models. LP readout is treated as the
primary evaluation because it matches the standard minimal-pair likelihood
protocol; the alternatives test whether explicit grammaticality
framing changes the apparent difficulty of \textsc{FreqBLiMP}. Exact prompt strings, the
aggregation procedure, and compute details are given in
Appendix~\ref{app:scoring-statistics}.

\section{Results and Discussion}
\label{sec:results-discussion}

\subsection{Sentence Likelihood Decreases with Lexical Rarity}
\begin{figure}[t]
  \centering
  \includegraphics[width=\columnwidth]{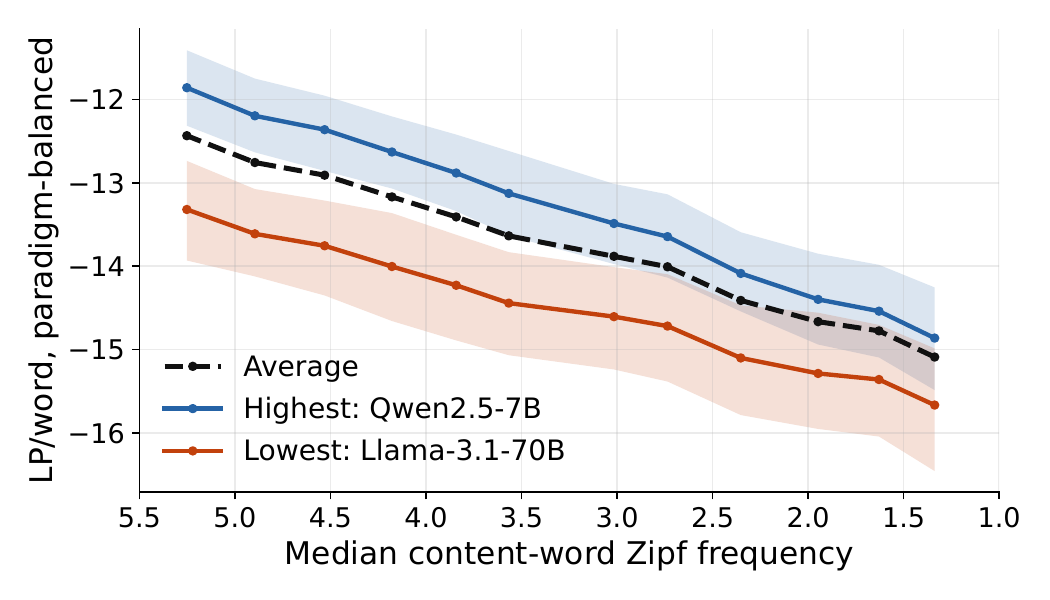}
  \caption{
  Paradigm-balanced word-normalized log probability of grammatical sentences
  as a function of realized Zipf frequency. The dashed black line shows the
  base-model average; blue and orange show the base models with the highest and
  lowest mean word-normalized log probability across the plotted Zipf bins.
  Shaded bands indicate 95\% confidence intervals over paradigm means. The
  corresponding all-model plot is shown in
  Appendix~\ref{app:all-model-diagnostics}.
    }
  \label{fig:lp-zipf}
\end{figure}

We first investigate whether the
frequency manipulation affects model expectation. For each model, we compute word-normalized log probability for the grammatical
member of each minimal pair, normalizing by word count to control for sentence
length. We then use paradigm-balanced binning: item scores are first averaged
within each model--paradigm--Zipf-bin group, and these paradigm means are then
averaged equally within each Zipf bin. Each plotted bin retains
broad paradigm coverage, with at least $58$ of $67$ paradigms represented.

Figure~\ref{fig:lp-zipf} shows that lower realized Zipf frequency corresponds
to lower sentence likelihood across all base models. The trend is monotonic for
each model, with Spearman $\rho=.979$--$.993$, and word-normalized likelihood
decreases by $2.58$ points on average from the highest to the lowest Zipf bin,
with a model range of $2.20$--$2.89$. This confirms that the frequency manipulation
affects model expectation, even though Zipf values are computed from an
external corpus-based resource rather than from model-specific training counts.

\subsection{Grammatical Preferences Are Robust on Aggregate}
\label{subsec:grammatical-performance}
\begin{figure}[t]
  \centering
  \includegraphics[width=\linewidth]{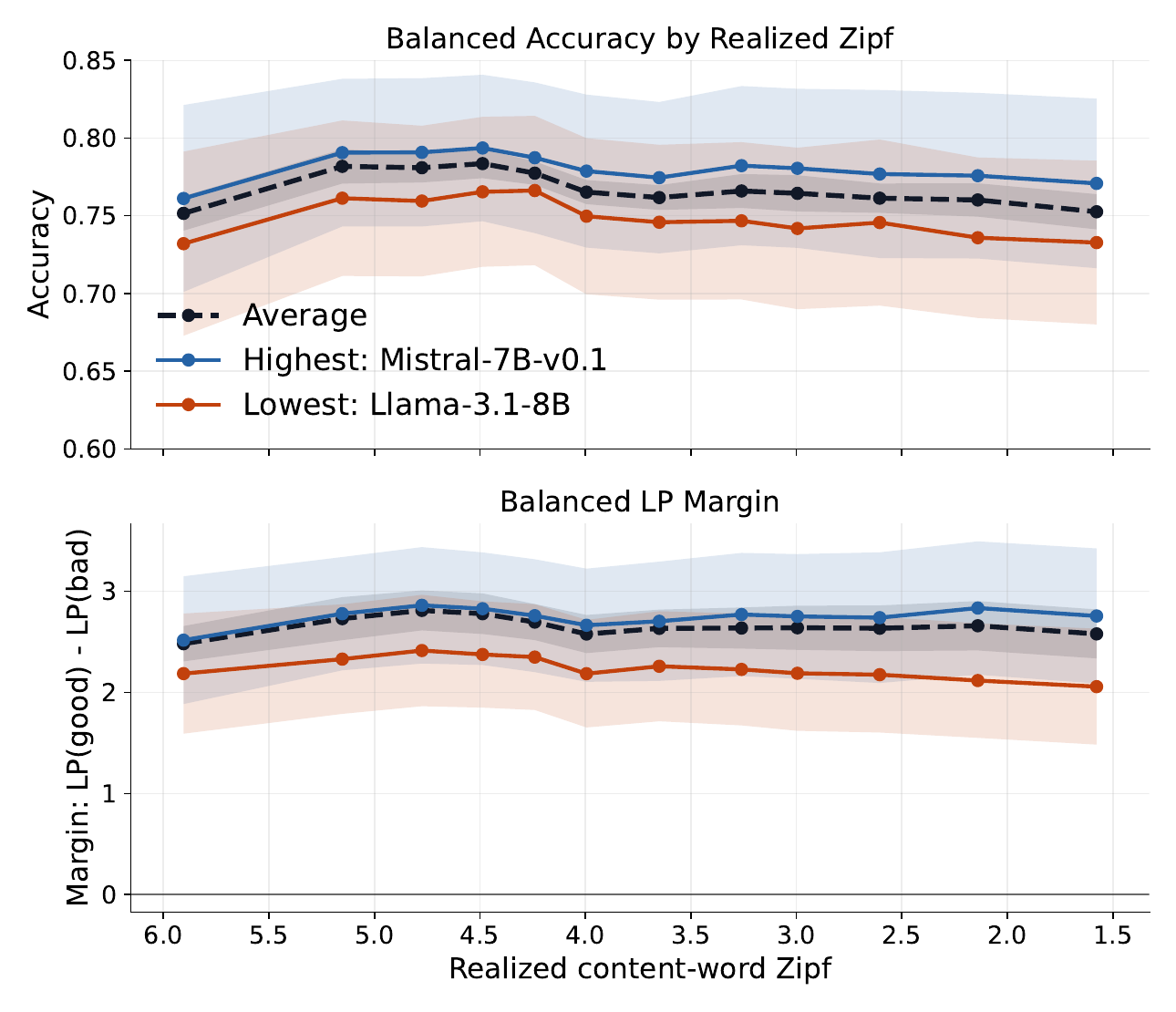}
  \caption{
  Paradigm-balanced contrastive accuracy and log-probability margin as a
  function of realized Zipf frequency. The dashed black line shows the
  base-model average; blue and orange show the base models with the highest and
  lowest mean contrastive accuracy across the plotted Zipf bins. Shaded bands indicate
  95\% confidence intervals over paradigm means.
  The corresponding all-model plot without confidence bands is shown in
  Appendix~\ref{app:all-model-diagnostics}.
}
  \label{fig:grammatical-performance-zipf}
\end{figure}

Having established that rare lexicalizations receive much lower sentence
likelihood, we next ask whether this likelihood penalty also weakens
BLiMP-style grammatical preferences.
Figure~\ref{fig:grammatical-performance-zipf} shows that contrastive
grammatical performance is robust on aggregate.\footnote{Original BLiMP shows the same qualitative likelihood--accuracy
dissociation (Figure~\ref{fig:original-zipf-diagnostics}).} Accuracy declines only modestly
as realized Zipf frequency decreases. The $\Delta$LP also declines on average,
but much less sharply than absolute sentence likelihood in
Figure~\ref{fig:lp-zipf}, and it remains positive even in the lowest-frequency
bins. Thus, rare words make sentences less expected in absolute terms, but they
do not overall reverse the model's preference for the grammatical member of the
pair.

Table~\ref{tab:generated-regime-performance} summarizes the head--extreme-tail
changes by model. Most models lose only a few accuracy points, with small
changes in $\Delta$LP.
Model-wise analysis revealed that Qwen2.5-7B showed the smallest accuracy gap in head and extreme-tail; in contrast, Gemma-4-31B exhibited the largest drop. 
The other models fall between these cases. 
These results confirm that grammatical preferences are robust in the aggregate, while frequency sensitivity is not a uniform consequence of lexical rarity, nor does it scale monotonically with model size.


\begin{table}[t]
  \centering
  \footnotesize
  \begin{tabular}{lrrrrr}
    \hline
    \textbf{Model} & \textbf{H} & \textbf{T} & \textbf{XT}
    & $\Delta_{\mathrm{XT-H}}$ & \textbf{Mrg. $\Delta$} \\
    \hline
    G4-E4B     & 78.3 & 77.2 & 76.4 & -1.9 & -0.08 \\
    G4-31B     & 78.6 & 77.5 & 76.0 & -2.6 & -0.24 \\
    L3.1-8B    & 75.7 & 75.1 & 73.9 & -1.9 & -0.21 \\
    L3.1-70B   & 75.9 & 75.3 & 74.5 & -1.4 & -0.08 \\
    M-7B       & 78.5 & 78.1 & 77.7 & -0.8 &  0.07 \\
    Q2.5-7B    & 78.1 & 78.2 & 77.5 & -0.6 &  0.10 \\
    Q2.5-72B   & 77.3 & 76.9 & 75.7 & -1.6 & -0.10 \\
    \hline
  \end{tabular}
  \caption{
    Base-model accuracy under LP readout across generated regimes.
        $\Delta_{\mathrm{XT-H}} =
        \mathrm{Acc}_{\mathrm{XTail}}-\mathrm{Acc}_{\mathrm{Head}}$, in percentage
        points.
  }
  \label{tab:generated-regime-performance}
\end{table}

The aggregate result leaves open where the remaining losses come from\footnote{The aggregate Head-to-XTail decline also persists after matching items on
model-specific tokenized length (Appendix~\ref{app:length-matching}).}. We
therefore next break the head--extreme-tail change down by linguistic domain
and paradigm.

\subsection{Frequency Effects Concentrate in Contrasts Requiring
Lemma-Specific Information}
\label{subsec:phenomenon-effects}

\begin{figure*}[t]
\centering
\includegraphics[width=0.49\textwidth]{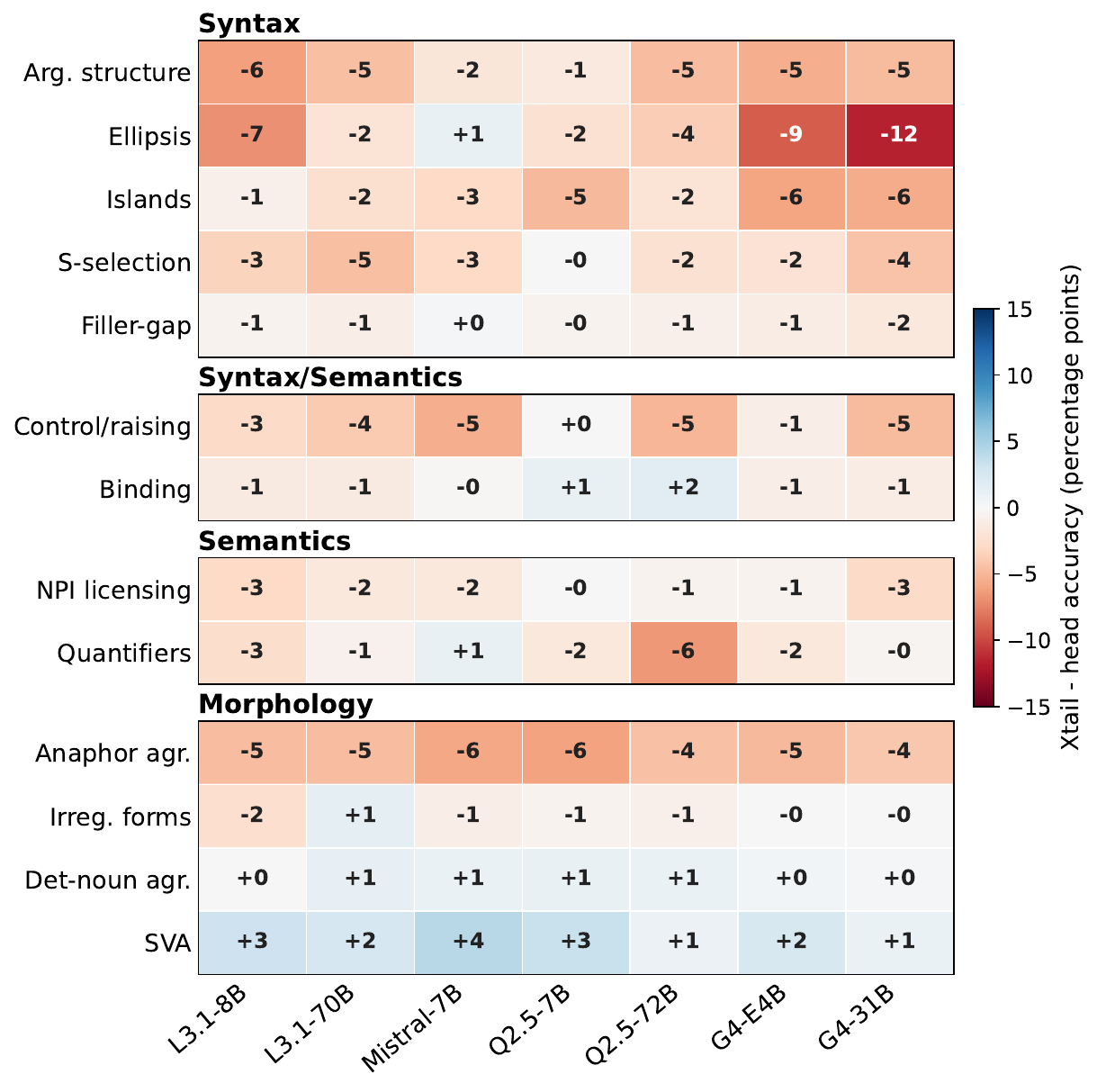}
\hfill
\includegraphics[width=0.49\textwidth]{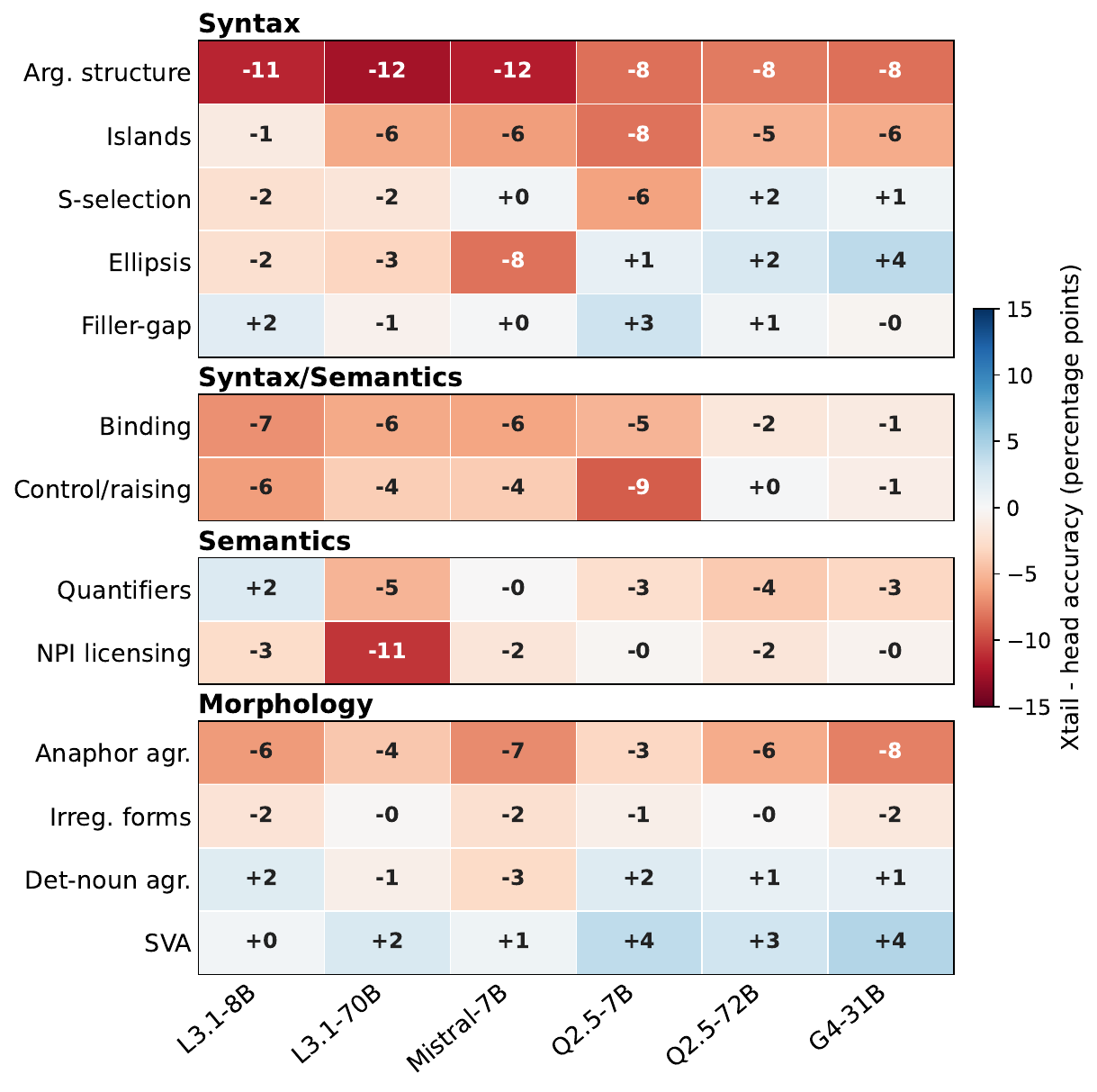}
\caption{Head--extreme-tail accuracy change by BLiMP phenomenon and base model
under LP readout (left) and Yes/No (right). Rows are grouped by
linguistic field. Negative values indicate lower accuracy under extreme-tail
lexicalization; positive values indicate higher accuracy. The LP panel includes
all seven base models. Gemma-4-E4B is excluded from the Yes/No panel because
that readout is near chance for this model.}
\label{fig:phenomenon-heatmap}
\end{figure*}

The aggregate results in Section~\ref{subsec:grammatical-performance} average
over heterogeneous grammatical phenomena. We therefore ask whether the
head--extreme-tail decline reflects a broad weakening of grammatical
discrimination, or whether it is concentrated in particular kinds of
contrasts. Figure~\ref{fig:phenomenon-heatmap} shows the head--extreme-tail
change in accuracy by phenomenon under the conventional LP readout and the
Yes/No readout, grouped by linguistic field. The pattern is highly non-uniform. Median field-level accuracy changes are
positive for morphology under both readouts ($+0.4$ pp) and negative elsewhere:
$-2.7$, $-1.3$, and $-2.8$ pp for syntax, syntax--semantics, and semantics
under LP, and $-4.0$, $-4.9$, and $-1.7$ pp under Yes/No. Thus, rare lexicalization does not uniformly hurt
grammar. The strongest effects appear where the contrast depends on
lemma-specific information, including verb argument structure as well as noun
properties such as gender and animacy. The weakest effects appear where the
contrast is carried by overt local morphosyntax. In the following, we use Yes/No to rank the
localized effects because it probes acceptability directly and is less coupled
than LP to lexical probability and other non-grammatical sentence properties.
Gemma-4-E4B is excluded from the Yes/No panel because its accuracy is already
close to chance in the head regime (56.4\%) and falls below chance in the
extreme tail (49.4\%).

The largest negative changes occur in paradigms that require lemma-specific
information, including binding and anaphor contrasts sensitive to gender and
animacy, as well as transitivity, passivization, and argument drop. BLiMP
already found models to be relatively strong on agreement
morphology but weaker on argument structure and other subtle syntactic or
semantic contrasts \citep{warstadt2020blimp}; \textsc{FreqBLiMP} shows that
this weakness becomes more pronounced as the relevant lexical items move into
the tail. As shown in Table~\ref{tab:localized-paradigms}, large mean Yes/No
decreases occur in \textit{principle A domain 3}, \textit{passive 1},
\textit{drop argument}, \textit{passive 2}, \textit{causative}, and
\textit{anaphor gender agreement}. The table also reports
\textit{expletive there subject raising}, but we interpret this control/raising
result cautiously because its verbs do not always fall within the
intended tail and extreme-tail Zipf windows, due to fallbacks from the limited pool. For example, an extreme-tail passive example such as
\textit{Splicers weren't \underline{readjusted}} vs.
*\textit{Splicers weren't \underline{flounced}} requires knowing that
\textit{readjust} can take an object and therefore be passivized, while
\textit{flounce} is normally intransitive in the intended frame. The largest
drops therefore reflect an interaction between lexical rarity and access to
lemma-specific grammatical information, rather than a general loss of
syntactic knowledge.

\begin{table}[t]
  \centering
  \setlength{\tabcolsep}{2.8pt}
  \resizebox{\columnwidth}{!}{%
  \begin{tabular}{@{}lrrrr@{}}
    \hline
    & \multicolumn{2}{c}{\textbf{LP}} & \multicolumn{2}{c}{\textbf{Yes/No}} \\
    \textbf{Paradigm} & \textbf{H} & \textbf{$\Delta \pm$ SD}
      & \textbf{H} & \textbf{$\Delta \pm$ SD} \\
    \hline
    \multicolumn{5}{@{}l}{\textit{Largest decreases}} \\
Principle A domain 3 & 65.3 & $-9.9 \pm 5.0$ & 90.5 & $-17.9 \pm 3.9$ \\
passive 1 & 79.9 & $-14.1 \pm 4.1$ & 91.4 & $-15.1 \pm 2.7$ \\
drop argument & 66.6 & $-1.6 \pm 2.4$ & 87.0 & $-14.4 \pm 5.0$ \\
passive 2 & 81.3 & $-9.6 \pm 2.8$ & 93.5 & $-13.5 \pm 1.0$ \\
expletive there subj. raising & 85.7 & $+2.5 \pm 5.2$ & 88.5 & $-12.4 \pm 12.0$ \\
causative & 73.3 & $-0.1 \pm 3.5$ & 93.1 & $-9.2 \pm 4.0$ \\
anaphor gender agr. & 97.4 & $-7.5 \pm 0.9$ & 97.9 & $-8.5 \pm 2.2$ \\
    \hline
    \multicolumn{5}{@{}l}{\textit{Largest increases}} \\
wh subj. gap, long & 95.7 & $-2.3 \pm 2.3$ & 75.5 & $+7.3 \pm 6.3$ \\
wh subj. gap & 94.2 & $+0.8 \pm 0.7$ & 80.2 & $+6.3 \pm 1.6$ \\
distractor agr., RC & 65.5 & $+9.1 \pm 5.2$ & 82.3 & $+5.0 \pm 4.0$ \\
    \hline
  \end{tabular}
  }
  \caption{Mean Head accuracy (H, \%) and mean $\pm$ SD of the
  XTail-minus-Head accuracy change ($\Delta$, percentage points) across the six
  base models shared by LP and Yes/No analysis. Paradigms are ordered within
  each group by mean Yes/No change. Full 67-paradigm results under both readouts are reported in
Appendix~\ref{app:phenomenon-tables}.}
  \label{tab:localized-paradigms}
\end{table}

The positive changes in Table~\ref{tab:localized-paradigms} appear to reflect
cases where rarity reduces misleading lexical or collocational cues, allowing
the model to rely more on overt morphosyntactic information. The two
subject-gap paradigms show the largest increases. Agreement paradigms also
illustrate this pattern clearly: their contrasts are supported by overt number
morphology, but generated head items can introduce lexical traps, as in
\textit{The man \underline{bags}} vs. *\textit{The man \underline{bag}}, or
strong local attractors near the verb. The smaller positive change for \textit{principle A c-command} likely reflects
a related effect: head items can contain animate distractors that make the
incorrect reflexive locally plausible, as in \textit{Some title that switches
Carol hosted \underline{itself}} vs.\ *\textit{Some title that switches Carol
hosted \underline{herself}}. Extreme-tail items are less often confounded in
this way.
Rare lexicalization also does not amplify relative-clause subject--verb
interference, and often reduces the measured attractor penalty
(Appendix~\ref{app:attractor-analysis}).

\subsection{Human Validation}
\label{subsec:human-validation}

To test whether declining model accuracy reflects weaker grammatical
generalization or ambiguous generated items, three native English speakers
judged 720 pairs: 15 per paradigm for 12 paradigms in each of the Original,
Head, Tail, and XTail conditions. Eight paradigms were selected because model
accuracy declined with frequency, three were stable controls, and one was
exploratory. Order and acceptable-sentence position were randomized per
annotator. Annotators selected one sentence or ``cannot decide,'' gave a reason
for non-decisions, and rated the intended acceptable sentence's naturalness
from 1 to 5. All three answered all 12 catch trials correctly.

Table~\ref{tab:human-validation} reports the results, with correct selection
computed over decided responses only. In the declining arm it falls from
$97.7\%$ in Head to $87.7\%$ in XTail while non-decisions rise from $15.6\%$ to
$32.5\%$, increasingly citing unknown words ($1.8\%$ to $53.8\%$); combining
the two, only about $59\%$ of XTail items were both decided and correct. The
stable arm holds near $98\%$ throughout despite comparable naturalness in XTail
($2.00$ vs.\ $2.27$), so naturalness loss alone does not predict which
contrasts degrade. Human and model accuracy therefore move in the same direction across the
two arms, declining together in the declining arm and holding in the stable
arm (Table~\ref{tab:human-validation}). Under two-of-three agreement, 593 of 720 pairs ($82.4\%$)
supported the intended key, and manual review of the 11 dissenting pairs did
not support reversing any.

Matching the annotated items to Yes/No scores, restricting to human-supported
items raises accuracy but does not remove the decline: the Head-to-XTail
decrease in the declining arm is $6.7$ points on all items and $4.2$ points
after filtering, while the stable arm shows no consistent decrease. The human
and model columns are computed over different response sets and are not
comparable in level (Appendix~\ref{app:human-validation}). Stimulus ambiguity
therefore contributes to, but does not fully explain, the measured degradation.

\begin{table}[t]
\centering
\setlength{\tabcolsep}{4pt}
\begin{tabular}{@{}lrrrrr@{}}
\hline
\textbf{Condition} & \textbf{Corr.} & \textbf{CD} & \textbf{Nat.}
& \textbf{Y/N} & \textbf{Y/N$_{\mathrm{f}}$} \\
\hline
\multicolumn{6}{@{}l}{\textit{Declining paradigms}} \\
\quad Original & 99.0 & 14.7 & 4.21 & --   & --   \\
\quad Head     & 97.7 & 15.6 & 3.28 & 90.0 & 92.6 \\
\quad Tail     & 91.5 & 21.4 & 2.67 & 84.5 & 88.3 \\
\quad XTail    & 87.7 & 32.5 & 2.27 & 83.3 & 88.4 \\
\multicolumn{6}{@{}l}{\textit{Stable paradigms}} \\
\quad Original & 97.7 &  5.2 & 3.90 & --   & --   \\
\quad Head     & 98.4 &  7.4 & 2.15 & 90.4 & 90.7 \\
\quad Tail     & 97.6 &  7.4 & 2.02 & 95.6 & 96.6 \\
\quad XTail    & 98.4 &  5.2 & 2.00 & 92.8 & 92.8 \\
\hline
\end{tabular}
\caption{Human validation by paradigm group and condition, over three
annotators. \textbf{Corr.}\ is the percentage of decided responses selecting
the intended acceptable sentence, \textbf{CD} the percentage of ``cannot
decide'' responses, \textbf{Nat.}\ mean naturalness (1--5) of the intended
acceptable sentence. \textbf{Y/N} is mean Yes/No accuracy (\%) on the same
items over six base and three instruction-tuned models, excluding G4-E4B;
\textbf{Y/N$_{\mathrm{f}}$} restricts to items with two-of-three human support.
Original items are outside the generated-regime model comparison.}
\label{tab:human-validation}
\end{table}

\subsection{Post-Training Makes Log-Probabilities Less Sensitive to Grammaticality}
\label{subsec:scoring-interface}

\begin{table}[t]
\centering
\footnotesize
\begin{tabular}{lrrrrrr}
\hline
& \multicolumn{2}{c}{\textbf{LP readout}}
& \multicolumn{2}{c}{\textbf{Temp. LP}}
& \multicolumn{2}{c}{\textbf{Yes/No}} \\
\textbf{Model} & \textbf{Acc.} & \textbf{$\Delta$}
& \textbf{Acc.} & \textbf{$\Delta$}
& \textbf{Acc.} & \textbf{$\Delta$} \\
\hline
\multicolumn{7}{l}{\textit{Base models}} \\
G4-E4B       & 77.3 & -1.9 & 78.8 & -1.2 & 52.9 & -7.0 \\
G4-31B       & 77.4 & -2.6 & 80.0 & -0.9 & 81.1 & -1.6 \\
L3.1-8B      & 74.9 & -1.9 & 79.2 & -1.5 & 79.5 & -2.7 \\
L3.1-70B     & 75.3 & -1.4 & 79.9 & -0.5 & 82.7 & -4.5 \\
M-7B         & 78.1 & -0.8 & 80.7 & -0.1 & 73.0 & -3.9 \\
Q2.5-7B      & 77.9 & -0.6 & 79.5 & -1.0 & 81.7 & -2.6 \\
Q2.5-72B     & 76.6 & -1.6 & 79.5 & -0.6 & 85.9 & -1.6 \\
\hline
\multicolumn{7}{l}{\textit{Instruction-tuned models}} \\
G4-31B-I     & 58.9 & -0.7 & 66.2 & +0.1 & 88.3 & -1.4 \\
L3.1-70B-I   & 76.4 & +1.7 & 81.1 & -0.6 & 85.1 & -4.5 \\
Q2.5-72B-I   & 75.2 & -1.6 & 78.0 & -0.6 & 87.2 & -1.2 \\
\hline
\end{tabular}
\caption{
Accuracy under different scoring interfaces. Acc. is mean accuracy across the
generated Head, Tail, and XTail regimes. $\Delta$ is XTail minus Head accuracy in percentage points. 
}
\label{tab:scoring-interface}
\end{table}


This section investigates how different scoring interfaces affect grammatical acceptability evaluation. 
Table~\ref{tab:scoring-interface} summarizes each method by its average accuracy
across generated regimes and its head--extreme-tail change. Per-regime accuracies and margins for every scoring method are reported
separately in Appendix~\ref{app:per-method-regimes}. For base models, Template LP is the most consistent intervention: it raises
average accuracy from $76.8\%$ to $79.7\%$ and reduces the average frequency gap
from $1.5$ points to $0.8$ points. Base-model Yes/No scoring averages $76.7\%$,
but is less uniform as expected. It fails for G4-E4B and underperforms LP readout
for M-7B, while giving large gains for larger models such as G4-31B, L3.1-70B,
and Q2.5-72B; within the Gemma, Llama, and Qwen families, the larger base model
obtains the higher Yes/No accuracy. Thus, explicit acceptability readout can
expose grammatical knowledge and frequency effects that LP readout misses.
Yes/No reveals large Head-to-XTail declines in \emph{drop argument} and
\emph{causative} that are nearly absent under LP readout, where Head accuracy is already about 20 points lower. Reverse cases also occur, such as \emph{regular plural SVA 2}, which is stable
under Yes/No but declines under LP readout
(Appendix~\ref{app:scoring-interface-disagreements}). Yes/No nevertheless
depends on the model's ability to use the prompted metalinguistic interface and
does not reduce the average frequency gap relative to LP.

Instruction-tuned models show a sharper separation between sentence likelihood
and grammaticality judgment. G4-31B-I averages only $58.9\%$ under LP readout, but
reaches $88.3\%$ under Yes/No scoring, the highest generated-regime accuracy in the
table; L3.1-70B-I and Q2.5-72B-I show the same direction. 
Furthermore, comparisons against the corresponding base models show that instruction tuning consistently reduces LP-readout accuracy.
These results suggest that
post-training can make grammaticality judgments more accessible through a
prompted interface, while making raw LP readout less directly interpretable:
the probability of a bare sentence may reflect alignment and dialogue-format
preferences rather than grammatical acceptability. 
The gains from instruction tuning also depend on model scale. 
An additional experiment showed that Qwen2.5-7B-Instruct improves only modestly, from $73.9\%$ under LP readout to $78.7\%$ under Yes/No scoring.
We therefore treat LP,
Template LP, and Yes/No as complementary probes of implicit sentence likelihood
and explicit metalinguistic judgment.

\subsection{Collocational Support Does Not Explain the Frequency Effects}
\label{subsec:original-head-gap}

Rarer vocabulary changes not only frequency but also semantic plausibility,
selectional fit, and constructional naturalness, the concern motivating
``colorless green'' evaluations \citep{gulordava2018colorless,
chomsky1957syntactic}. We use COCA \citep{davies2008coca} as an external proxy for local
collocational support: for each minimal pair we extract local content-word
lemma dependencies with spaCy \citep{honnibal2020spacy} (e.g., verb--object
pairs) and compute the mean $\log(1+\mathrm{doc\ count})$ over them.

\begin{table}[t]
\centering
\setlength{\tabcolsep}{4.0pt}
\begin{tabular}{lrrrr}
\hline
\textbf{Dataset} & \textbf{Support} & \textbf{Zero pairs} & \textbf{Acc.} & \textbf{$\Delta$LP} \\
\hline
BLiMP        & 4.33 & 9.3 & 79.7 & 3.53 \\
Head         & 2.59 & 17.7 & 77.5 & 2.70 \\
Tail         & 0.56 & 71.3 & 76.9 & 2.67 \\
XTail        & 0.25 & 86.3 & 76.0 & 2.62 \\
\hline
\end{tabular}
\caption{
COCA local-collocational support by dataset. Support is mean
$\log(1+\mathrm{doc\ count})$ over extracted local lemma pairs; zero pairs is
the percentage of extracted pairs unattested in COCA. Acc. and $\Delta$LP are means over item--model rows for the seven base models.
}
\label{tab:local-support-by-dataset}
\end{table}

Table~\ref{tab:local-support-by-dataset} shows that the frequency manipulation
removes local collocational support almost entirely: mean support falls from
$4.33$ in original BLiMP to $0.25$ in XTail, and the proportion of extracted
lemma pairs unattested in COCA rises from $9.3\%$ to $86.3\%$. Aggregate
LP-readout accuracy over the same range falls by $3.7$ points, and by only
$1.5$ points within the generated regimes. An almost total collapse of local
collocational support therefore costs little aggregate contrastive accuracy,
which is consistent with the aggregate robustness reported in
Section~\ref{subsec:grammatical-performance}.

\begin{table}[t]
\centering

\begin{tabular}{lrrr}
\hline
\textbf{Support quartile} & \textbf{Support} & \textbf{Acc.} & \textbf{$\Delta$LP} \\
\hline
Lowest  & 0.33 & 75.8 & 2.56 \\
Q2      & 0.77 & 76.7 & 2.65 \\
Q3      & 1.24 & 77.4 & 2.71 \\
Highest & 2.19 & 78.0 & 2.81 \\
\hline
\end{tabular}
\caption{
LP-readout performance by local-support quartile. Quartiles are assigned within
each paradigm and generated regime, then averaged equally across regimes.
Support is mean $\log(1+\mathrm{doc\ count})$; Acc. and $\Delta$LP are base-model means.
}
\label{tab:support-quartiles}
\end{table}

Section~\ref{subsec:human-validation} addressed naturalness directly for twelve
paradigms. As a corpus-side measure across all 67, we assign COCA support
quartiles within each paradigm and regime, so that compared items share both a
construction and a Zipf window. Accuracy rises monotonically across quartiles
(Table~\ref{tab:support-quartiles}), a gradient consistent in direction but
small: a spread of $2.2$ accuracy points, versus paradigm-level
head--extreme-tail changes an order of magnitude larger
(Table~\ref{tab:localized-paradigms}). Local collocational support is a
real but bounded covariate, not a sufficient account of the effects in
Section~\ref{subsec:phenomenon-effects}.

Comparing original BLiMP with Head isolates regeneration itself, since the two
share high-frequency vocabulary but not the generation pipeline. Head is lower
for six of seven base models, with a median drop of $1.6$ points, concentrated
in a small number of filler--gap, island, and tough/raising paradigms rather
than spread across the benchmark (Appendix~\ref{app:original-head-gap}). These
are largely disjoint from the paradigms that decline under increasing lexical
rarity, indicating that the regeneration gap and the frequency gradient are
separable effects.

Comparisons among the \textsc{FreqBLiMP} regimes hold the generation pipeline
fixed while varying only the target Zipf window, and are therefore the more
reliable basis for interpreting frequency effects.

\section{Conclusion and Future Work}
\label{sec:conclusion}

We introduced \textsc{FreqBLiMP}, a frequency-controlled extension of BLiMP for
testing whether LLMs preserve grammatical preferences when content words are
shifted into the lexical long tail. 
Our findings highlight lexical frequency as a useful control for grammatical evaluation benchmarks.


Future work should connect frequency control more directly to model exposure by measuring item frequency in accessible pretraining corpora. 
Larger-scale human judgments could further disentangle grammatical degradation
from lexical oddity, semantic implausibility, and accidental ambiguity in
generated samples.
Finally, representation-level analyses could test how rare lexicalization affects the encoding of grammatical contrasts, and cross-linguistic extensions could test whether the robustness observed here generalizes beyond English BLiMP-style templates.

\section*{Limitations}
\label{sec:limitations}

\textsc{FreqBLiMP} uses external Zipf estimates as proxies for lexical
familiarity rather than model-specific training counts. These estimates do not
directly measure exposure in each model's pretraining distribution, although our
likelihood diagnostics show that the lower-Zipf regimes consistently receive
lower model likelihoods.

A second limitation is residual noise from automatic generation under BLiMP's
heterogeneous paradigm constraints. Human validation confirms that this noise
is structured rather than uniform. In paradigms selected for declining model
performance, correct selection among decided responses fell from $97.7\%$ in
Head to $87.7\%$ in XTail, while non-decisions rose from $15.6\%$ to $32.5\%$.
Stable controls remained near $98\%$. Restricting evaluation to items supported
by at least two annotators reduced the Yes/No Head-to-XTail decline from $6.7$
to $4.2$ points rather than eliminating it. Residual noise therefore
contributes to the measured degradation but does not fully explain it. Representative ambiguous items from our qualitative audit are
given in Appendix~\ref{app:qualitative-examples}.

The validation study is itself limited. It uses a stratified sample of 720
items from 12 of the 67 paradigms rather than a benchmark-wide random sample,
and conditions contain independently sampled sentences rather than matched
lexicalizations. It should therefore be interpreted as a stimulus-quality audit,
not as a controlled estimate of the human frequency effect. In addition, some
paradigms use curated fallbacks when the target frequency window cannot be
filled, particularly in control and raising. We consequently interpret those
paradigms cautiously.

Finally, our experiments are limited to English, BLiMP-style template
paradigms, the evaluated open-weight LLMs, and the scoring interfaces considered
here. The results should therefore be interpreted as evidence about this
controlled setting, not as a complete account of grammatical generalization
across languages, models, or evaluation formats.

\section*{Ethical Considerations}
\textsc{FreqBLiMP} consists of synthetic English minimal pairs generated from
BLiMP-style templates. It does not contain user data or intentionally identify
real individuals. However, because the lexical overlay draws from broad lexical
and knowledge-base resources, generated examples may occasionally contain
borderline, slang, vulgar, or otherwise sensitive lexical material, or combine
words in awkward ways. We filter such material where possible and manually
inspect generated samples during development, but some borderline cases may
remain. The dataset is intended for research evaluation of language models, not
for user-facing text generation or normative claims about acceptability. Licenses and terms of use for all datasets, lexical resources, software, and
models used in this work are listed in Appendix~\ref{app:licenses}.

\section*{Acknowledgment}
This work was supported by JSPS KAKENHI Grant Number JP25K03233 and JP25H01149.

\bibliography{custom}

\appendix
\makeatletter
\@addtoreset{table}{section}
\@addtoreset{figure}{section}
\makeatother
\renewcommand{\thetable}{\Alph{section}\arabic{table}}
\renewcommand{\thefigure}{\Alph{section}\arabic{figure}}

\section{Lexical Overlay Implementation}
\label{app:lexical-overlay-implementation}

Section~\ref{subsec:lexical-overlay} summarizes the lexical overlay at a high
level. Here we give the implementation details needed to understand how nouns,
adjectives, and verbs are admitted into BLiMP-compatible templates without
turning generation into unconstrained lexical substitution.

\paragraph{Nouns.}
Noun candidates are drawn from Open English WordNet
\citep{mccrae-etal-2019-english}. We filter out forms poorly suited to BLiMP
templates, including duplicates, hyphenated forms, proper-name-like entries,
plural-looking singulars, language or demonym artifacts, acronyms, and entries
marked as vulgar, slang, or otherwise unsuitable where such metadata is
available. We also use manual blocklists for borderline cases observed during
development. Accepted nouns are mapped to BLiMP-compatible bundles using
WordNet metadata, definition-based heuristics, and Wikidata-derived person and
gender information for human-denoting nouns
\citep{vrandecic-krotzsch-2014-wikidata}. These signals are used to assign
animacy, countability, physicality, gender compatibility, and broad semantic
class. Polysemous nouns are rejected when their senses are dispersed across
incompatible classes; accepted nouns inherit matching BLiMP noun templates.

\paragraph{Adjectives.}
Adjective candidates are also drawn from Open English WordNet. We remove
relational-only and otherwise unstable adjectives, then map the remaining items
to compatibility bundles such as generic, animate-compatible, physical,
document-compatible, cleanable, and food-related adjectives. Each accepted
adjective inherits a template whose selectional restrictions match the assigned
bundle. Paradigms requiring specific adjectival behavior, such as control,
raising, and tough-adjective constructions, use curated predicate lists.

\paragraph{Verbs.}
For verbs, we construct a frame-annotated inventory with coarse
subcategorization labels such as intransitive, transitive, ditransitive,
prepositional, and particle frames. Frame labels are assigned and validated
using VerbNet \citep{kipper_schuler_2005_verbnet}, WordNet, corpus checks, and
manual review. Accepted verb lemmas are inserted into BLiMP-compatible verb
templates, with inflected forms generated automatically. Paradigms that depend
on more specific verbal behavior, such as control, raising, passivization,
causative--inchoative alternations, and object drop, use curated verb lists
rather than unconstrained frequency-filtered sampling.

The same minimum-pool fallback used during generation pads a curated list with
the nearest-frequency eligible predicates when fewer than ten candidates fall
inside the requested window. Table~\ref{tab:appendix-lexical-licensing} audits
the resulting frequency control at the critical contrast item itself.
Each percentage is computed over all $1{,}000$ items on the acceptable (G) or
unacceptable (B) side of each regime. The argument-structure predicates and
animate-selection nouns remain fully controlled, whereas the limited raising
and tough-predicate inventories often require fallback.

\begin{table}[t]
\centering
\resizebox{\columnwidth}{!}{%
\begin{tabular}{lrrrr}
\hline
& \multicolumn{2}{c}{\textbf{Tail in window}} &
  \multicolumn{2}{c}{\textbf{XTail in window}} \\
\textbf{Paradigm} & \textbf{G} & \textbf{B} & \textbf{G} & \textbf{B} \\
\hline
causative                         & 100.0 & 100.0 & 100.0 & 100.0 \\
inchoative                        & 100.0 & 100.0 & 100.0 & 100.0 \\
passive 1                         & 100.0 & 100.0 & 100.0 & 100.0 \\
passive 2                         & 100.0 & 100.0 & 100.0 & 100.0 \\
drop argument                     & 100.0 & 100.0 & 100.0 & 100.0 \\
animate subj. trans.              & 100.0 & 100.0 & 100.0 & 100.0 \\
animate subj. passive             & 100.0 & 100.0 & 100.0 & 100.0 \\
expletive there obj. raising      &  11.9 & 100.0 &   0.0 &  31.1 \\
expletive it obj. raising         &  12.6 & 100.0 &   0.0 &  32.6 \\
expletive there subj. raising     & 100.0 & 100.0 &   0.0 &  74.1 \\
tough vs. raising 1               & 100.0 &  10.7 &  10.5 &   0.0 \\
tough vs. raising 2               &  11.2 & 100.0 &   0.0 &  10.3 \\
\hline
\end{tabular}
}
\caption{Critical-item in-window coverage (\%) for paradigms using curated
lexical or predicate lists. G and B denote the acceptable and unacceptable sides
of the minimal pair. Rates are computed over all $1{,}000$ items per side and
regime.
}
\label{tab:appendix-lexical-licensing}
\end{table}

\section{Artifact Use and Licenses}
\label{app:licenses}

We use existing artifacts for research evaluation and lexical-resource construction. \textbf{Datasets and lexical resources:} BLiMP \citep{warstadt2020blimp} and Open English WordNet \citep{mccrae-etal-2019-english} are distributed under CC BY 4.0; Wikidata data \citep{vrandecic-krotzsch-2014-wikidata} is distributed under CC0; VerbNet \citep{kipper_schuler_2005_verbnet} is used under its stated VerbNet license; and COCA \citep{davies2008coca} is used under the English-Corpora.org access terms. COCA is used only as an external source of aggregate local collocational counts, and we do not redistribute COCA text. \textbf{Software:} spaCy \citep{honnibal2020spacy} and lemminflect \citep{jascob2019lemminflect} are distributed under the MIT License, and wordfreq code \citep{speer2022wordfreq} is distributed under the Apache License and includes redistributable frequency data under CC BY-SA 4.0.

\textbf{Models:} Mistral-7B-v0.1 \citep{jiang2023mistral} is distributed under Apache 2.0; Llama-3.1 \citep{grattafiori2024llama3} is distributed under the Llama 3.1 Community License; Gemma-4 \citep{gemmateam2026gemma4} models are distributed under the Gemma Terms of Use; and Qwen2.5 \citep{qwen2025qwen25} models are used under their corresponding Hugging Face license terms, with the 72B variants covered by the Qwen license rather than Apache 2.0. All models are used for inference-only evaluation, consistent with their intended research use.

\section{Dataset Diagnostics and Lexical Diversity}
\label{app:lexical-diversity}

Table~\ref{tab:appendix-regime-lexical-stats} gives the realized regime-level lexical statistics used in the diagnostics. Diversity is computed over frequency-controlled content-word occurrences. Table~\ref{tab:appendix-token-length} reports the corresponding sentence-length and tokenizer-length diagnostics, since rare lexicalization can increase subword tokenization even when word counts remain similar.

\begin{table*}[t]
\centering
\small
\begin{tabular}{lrrrrrrrr}
\hline
\textbf{Dataset / regime} & \textbf{Pairs} & \textbf{Mean Zipf} & \textbf{Median} & \textbf{IQR} & \textbf{Lemmas} & \textbf{Surfaces} & \textbf{Eff. lemmas} & \textbf{Top-20} \\
\hline
Original BLiMP & 67,000 & 4.26 & 4.26 & 0.65 & 2,607 & 2,808 & 441 & 14.3 \\
Head & 67,000 & 4.10 & 4.05 & 0.47 & 6,304 & 9,363 & 1,954 & 2.9 \\
Tail & 67,000 & 2.89 & 2.85 & 0.29 & 10,583 & 13,843 & 2,559 & 4.3 \\
XTail & 67,000 & 1.93 & 1.85 & 0.37 & 15,111 & 18,751 & 2,660 & 4.0 \\
\hline
\end{tabular}
\caption{
Regime-level dataset statistics. Mean Zipf, median Zipf, and IQR are computed
over realized frequency-controlled content-word occurrences; IQR is the
interquartile range. Eff. lemmas is the effective number of lemmas
($1/\sum_i p_i^2$), where $p_i$ is each lemma's share of frequency-controlled
content-word occurrences. Top-20 lemma share is reported as a percentage of
those occurrences.
}
\label{tab:appendix-regime-lexical-stats}
\end{table*}

\begin{table*}[t]
\centering
\small
\begin{tabular}{lrrr}
\hline
\textbf{Dataset / regime} & \textbf{Mean words} & \textbf{Mean tokens} & \textbf{Controlled content words} \\
\hline
Original BLiMP & 7.72 & 9.98 & -- \\
Head & 7.84 & 9.73 & 3.43 \\
Tail & 7.95 & 12.79 & 3.37 \\
XTail & 7.86 & 14.00 & 3.31 \\
\hline
\end{tabular}
\caption{Length and tokenization diagnostics for grammatical sentences. Token counts are averaged across the seven base-model tokenizers under LP readout; controlled content words are frequency-controlled lexical slots.}
\label{tab:appendix-token-length}
\end{table*}

\section{Scoring and Implementation Details}
\label{app:scoring-statistics}

\subsection{Exact Scoring Templates}

For Template LP, each sentence is embedded in the following string:
\begin{quote}\ttfamily
The following sentence is grammatically acceptable.\\
\\
\{sentence\}
\end{quote}

For Yes/No scoring with base models, the completion prompt is:
\begin{quote}\ttfamily
Your task is to evaluate the quality of given text.\\
Is the following sentence grammatically acceptable?\\
\\
\{sentence\}\\
Respond with Yes or No as your answer. Answer:
\end{quote}

For chat models, the system message is \texttt{Your task is to evaluate the quality of given text.} The user message is:
\begin{quote}\ttfamily
Is the following sentence grammatically acceptable? Respond with Yes or No as your answer.\\
\\
\{sentence\}
\end{quote}
The tokenizer chat template is then applied with generation prompting enabled. Yes/No scores use the next-token probability mass assigned to single-token variants \texttt{Yes}, \texttt{ Yes}, \texttt{\textbackslash nYes} and \texttt{No}, \texttt{ No}, \texttt{\textbackslash nNo}. The score is $p_Y/(p_Y+p_N)$.


\paragraph{Batched scoring.}
All reported scores use right-padded batches with padding excluded by the
attention mask. Position IDs are derived from this mask so that real tokens
retain contiguous positions independent of padding. Logits are converted to
\texttt{float32} before computing token probabilities and LP, while model
inference uses the configured model dtype.

\subsection{Statistical Aggregation}

For each model, paradigm, and bin or regime, we first average item-level scores within the paradigm and then average paradigm means equally. Confidence intervals are normal-approximation intervals over paradigm means:
\[
\bar{x} \pm 1.96 \frac{s}{\sqrt{n}},
\]
where $n$ is the number of represented paradigms. Reported aggregate results are paradigm-balanced rather than item-balanced.

\subsection{Compute Details}

Table~\ref{tab:implementation-versions} lists the main direct packages used for
dataset construction, preprocessing, evaluation, and analysis. The final
\textsc{FreqBLiMP} generation pipeline uses \texttt{wordfreq} for Zipf
frequencies and \texttt{lemminflect} for inflection; spaCy is used only in the
COCA local-support analysis.

\begin{table}[t]
\centering
\footnotesize
\begin{tabular}{ll}
\hline
\textbf{Package} & \textbf{Version} \\
\hline
\texttt{transformers} & 4.57.3 \\
\texttt{torch} & 2.8.0 \\
\texttt{wordfreq} & 3.1.1 \\
\texttt{lemminflect} & 0.2.3 \\
\texttt{spaCy} & 3.7.4 \\
\texttt{en\_core\_web\_sm} & 3.7.1 \\
\texttt{wn} & 0.13.0 \\
\texttt{datasets} & 4.4.1 \\
\texttt{pandas} & 2.3.3 \\
\texttt{numpy} & 2.0.2 / 1.26.4 \\
\texttt{scipy} & 1.13.1 \\
\hline
\end{tabular}
\caption{Main direct package versions used in dataset construction, preprocessing, evaluation, and analysis. \texttt{numpy}'s two versions indicate separate construction and evaluation environments.}
\label{tab:implementation-versions}
\end{table}

Zipf frequencies are computed with
\texttt{wordfreq.zipf\_frequency(text, "en")}. Inflection uses
\texttt{lemminflect} \citep{jascob2019lemminflect} with Penn-style tags such as \texttt{NN}/\texttt{NNS},
\texttt{JJ}/\texttt{JJR}/\texttt{JJS}, and
\texttt{VB}/\texttt{VBP}/\texttt{VBZ}/\texttt{VBG}/\texttt{VBD}/\texttt{VBN}.
The COCA local-support analysis uses spaCy dependency parses and lemmas to
extract local content-word dependencies.

Evaluations were run as inference-only jobs on NVIDIA H100 GPUs.
Models in the 4B--8B range used one H100, 31B models used two H100s, and
70B--72B models used four H100s. The final selected evaluation suite corresponds
to approximately 320 H100 GPU-hours. Auxiliary pilot runs not used in the paper are excluded. Dataset
generation used CPU jobs and did not involve model training.

\section{Additional Aggregate Results and Robustness Checks}
\label{app:aggregate-results}

\subsection{All-model Frequency Trends}
\label{app:all-model-diagnostics}

\begin{figure}[t]
  \centering
  \includegraphics[width=\columnwidth]{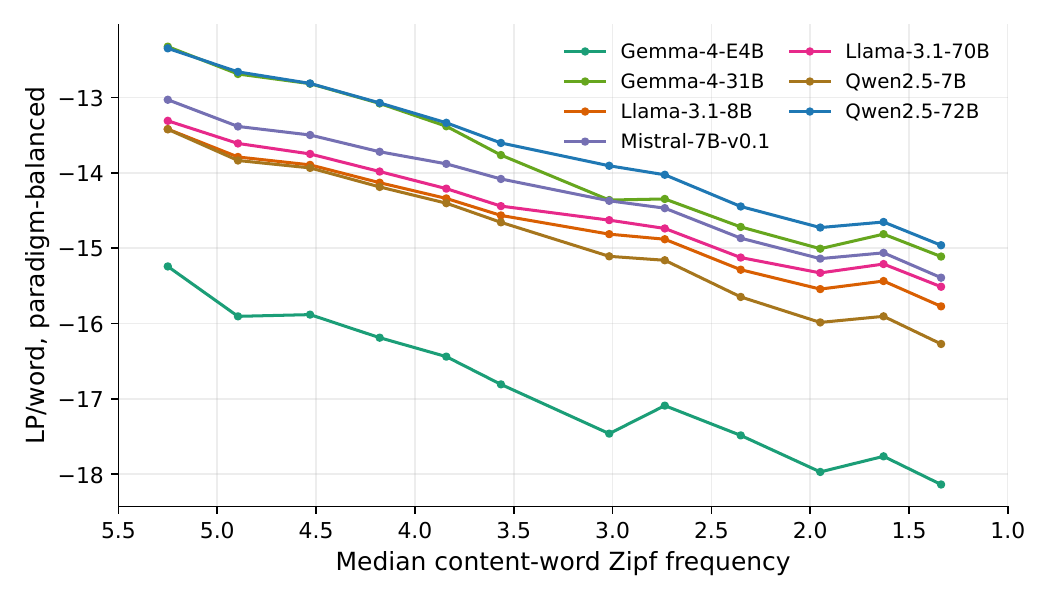}
  \caption{
    Paradigm-balanced word-normalized log probability of grammatical sentences
    as a function of realized Zipf frequency for all base models. Confidence
    bands are omitted for readability.
  }
  \label{fig:lp-zipf-all-models}
\end{figure}

\begin{figure}[t]
  \centering
  \includegraphics[width=\columnwidth]{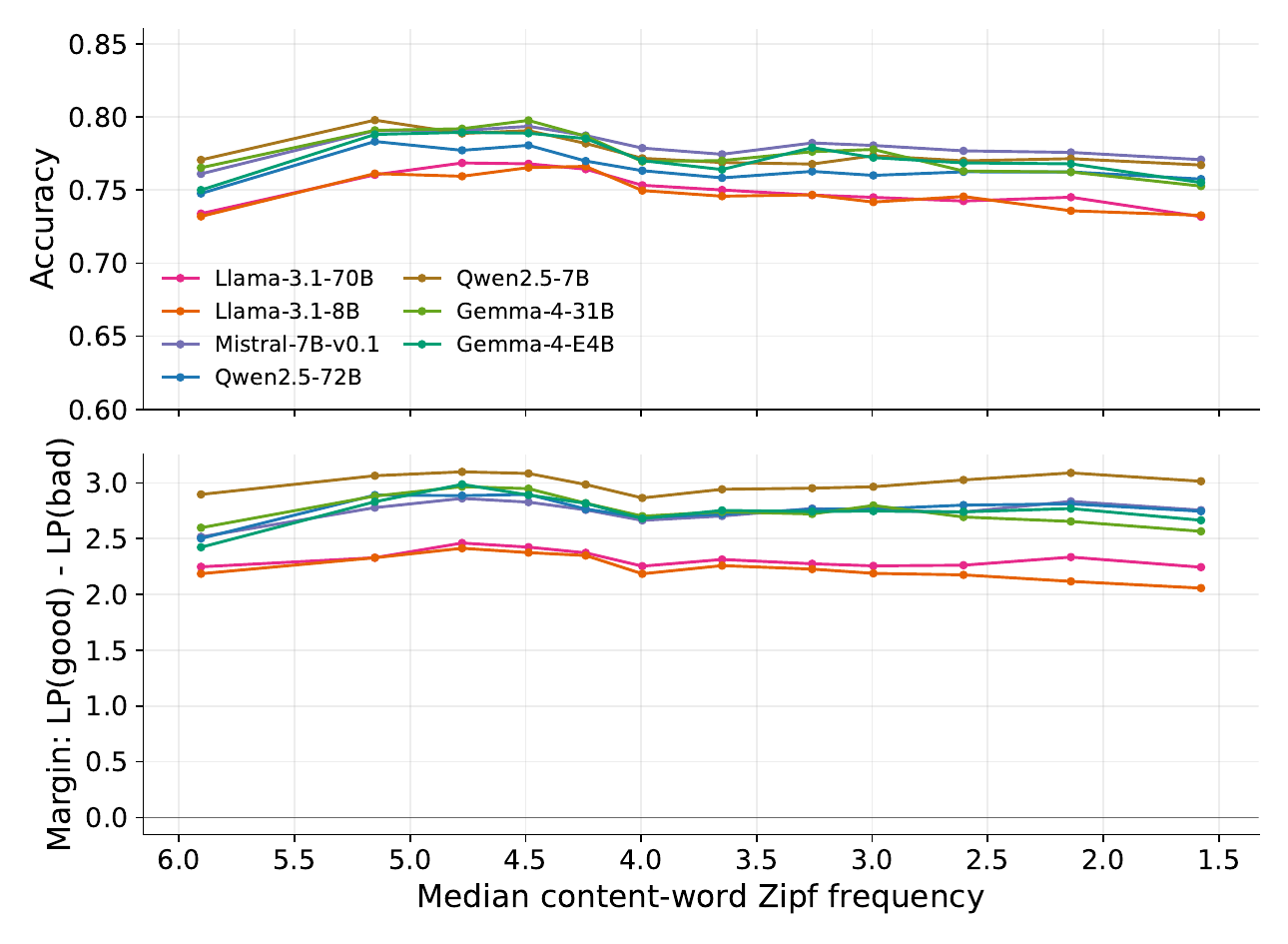}
  \caption{
    Paradigm-balanced contrastive accuracy and log-probability margin as a
    function of realized Zipf frequency for all base models. Confidence bands
    are omitted for readability.
  }
  \label{fig:performance-zipf-all-models}
\end{figure}

The main-text frequency plots show the base-model average and the two models
with the smallest and largest head-to-tail changes, in order to make the trends
readable with confidence intervals. Figure~\ref{fig:lp-zipf-all-models}
 shows the corresponding all-model
version without confidence bands, and Figure~\ref{fig:performance-zipf-all-models} the corresponding accuracy and margin plot. 

\subsection{Full Performance Results Per Method}
\label{app:per-method-regimes}

Table~\ref{tab:appendix-generated-method-performance} expands the scoring-interface summary by showing each generated regime separately. Margins are LP differences for LP and Template LP, and $P_{yes}(g)-P_{yes}(b)$ for Yes/No.

\begin{table*}[t]
\centering
\begin{tabular}{lrrrrrr}
\hline
\textbf{Model} & \textbf{H acc.} & \textbf{T acc.} & \textbf{XT acc.} & \textbf{H mrg.} & \textbf{T mrg.} & \textbf{XT mrg.} \\
\hline
\multicolumn{7}{l}{\textit{LP readout}} \\
G4-E4B & 78.3 & 77.2 & 76.4 & 2.83 & 2.73 & 2.74 \\
G4-31B & 78.6 & 77.5 & 76.0 & 2.86 & 2.77 & 2.62 \\
L3.1-8B & 75.7 & 75.1 & 73.9 & 2.34 & 2.26 & 2.12 \\
L3.1-70B & 75.9 & 75.3 & 74.5 & 2.36 & 2.33 & 2.28 \\
M-7B & 78.5 & 78.1 & 77.7 & 2.73 & 2.76 & 2.81 \\
Q2.5-7B & 78.1 & 78.2 & 77.5 & 2.95 & 3.02 & 3.05 \\
Q2.5-72B & 77.3 & 76.9 & 75.7 & 2.81 & 2.80 & 2.71 \\
G4-31B-I & 59.1 & 59.2 & 58.3 & 1.95 & 2.45 & 2.58 \\
L3.1-70B-I & 75.3 & 76.7 & 77.1 & 2.16 & 2.32 & 2.39 \\
Q2.5-72B-I & 75.9 & 75.4 & 74.3 & 2.98 & 3.09 & 3.03 \\
\hline
\multicolumn{7}{l}{\textit{Template LP}} \\
G4-E4B & 79.4 & 78.9 & 78.2 & 2.82 & 2.80 & 2.78 \\
G4-31B & 80.3 & 80.2 & 79.4 & 2.98 & 2.96 & 2.90 \\
L3.1-8B & 79.9 & 79.3 & 78.3 & 2.84 & 2.71 & 2.63 \\
L3.1-70B & 80.0 & 80.1 & 79.5 & 2.81 & 2.86 & 2.80 \\
M-7B & 80.7 & 80.8 & 80.6 & 2.92 & 3.02 & 3.04 \\
Q2.5-7B & 80.1 & 79.5 & 79.0 & 2.99 & 2.98 & 3.02 \\
Q2.5-72B & 79.7 & 79.8 & 79.1 & 2.92 & 2.97 & 2.94 \\
G4-31B-I & 66.1 & 66.4 & 66.2 & 3.60 & 4.48 & 4.86 \\
L3.1-70B-I & 81.3 & 81.1 & 80.8 & 3.04 & 3.12 & 3.13 \\
Q2.5-72B-I & 78.1 & 78.3 & 77.5 & 3.34 & 3.55 & 3.54 \\
\hline
\multicolumn{7}{l}{\textit{Yes/No}} \\
G4-E4B & 56.4 & 52.8 & 49.4 & 0.01 & 0.00 & 0.00 \\
G4-31B & 81.9 & 81.2 & 80.3 & 0.08 & 0.06 & 0.06 \\
L3.1-8B & 80.7 & 79.7 & 78.1 & 0.05 & 0.04 & 0.03 \\
L3.1-70B & 85.0 & 82.6 & 80.5 & 0.13 & 0.10 & 0.08 \\
M-7B & 74.9 & 73.3 & 71.0 & 0.03 & 0.02 & 0.02 \\
Q2.5-7B & 82.9 & 81.9 & 80.3 & 0.15 & 0.13 & 0.11 \\
Q2.5-72B & 86.6 & 86.1 & 84.9 & 0.12 & 0.11 & 0.09 \\
G4-31B-I & 89.0 & 88.3 & 87.6 & 0.41 & 0.45 & 0.44 \\
L3.1-70B-I & 87.2 & 85.3 & 82.7 & 0.33 & 0.35 & 0.31 \\
Q2.5-72B-I & 87.6 & 87.5 & 86.4 & 0.19 & 0.21 & 0.22 \\
\hline
\end{tabular}
\caption{Generated-regime performance grouped by scoring method. Accuracy columns are percentages.}
\label{tab:appendix-generated-method-performance}
\end{table*}

\subsection{Paradigm-Level Scoring-Interface Disagreements}
\label{app:scoring-interface-disagreements}

Scoring interfaces can yield qualitatively different frequency trajectories.
Across the six base models shared by both analyses, LP-readout accuracy on
\emph{drop argument} changes only from $66.6\%$ in Head to $65.0\%$ in XTail,
whereas Yes/No falls from $87.0\%$ to $72.6\%$. Similarly, LP-readout accuracy
on \emph{causative} changes from $73.3\%$ to $73.2\%$, while Yes/No falls from
$93.1\%$ to $83.9\%$. LP readout therefore starts about 20 points lower in Head
for both paradigms. Its nearly flat trajectory must be interpreted against these
lower baselines and does not by itself indicate greater stability.

The reverse pattern also occurs. For \emph{regular plural SVA 2}, LP-readout
accuracy falls from $91.1\%$ in Head to $87.4\%$ in XTail, whereas Yes/No remains
high and stable, changing from $95.8\%$ to $96.7\%$. A more dramatic
model-specific example is G4-31B on \emph{ellipsis N-bar 1}: LP-readout accuracy
falls from $60.9\%$ to $37.5\%$, while Yes/No remains stable at $92.6\%$ and
$92.2\%$. Thus, an individual interface can obscure either degradation or
stability, although the clearest aggregate disagreements more often involve
Yes/No recovering degradation missed by LP readout.

\subsection{Token-Length-Matched Analysis}
\label{app:length-matching}

Although Head and XTail sentences have similar word counts, XTail sentences
contain more subword tokens (Table~\ref{tab:appendix-token-length}). We therefore
matched Head and XTail items one-to-one without replacement within each model
and paradigm, using the mean token count of the two sentences in each pair.
Exact matches were preferred, with a maximum difference of one token.

\begin{table}[t]
\centering
\setlength{\tabcolsep}{4pt}
\begin{tabular}{lrrr}
\hline
\textbf{Model} & \textbf{Full $\Delta$} & \textbf{Matched $\Delta$}
& \textbf{Retained} \\
\hline
G4-E4B    & -1.9 & -3.3 & 38.3 \\
G4-31B    & -2.6 & -3.4 & 38.3 \\
L3.1-8B   & -1.9 & -2.7 & 27.7 \\
L3.1-70B  & -1.4 & -2.8 & 27.7 \\
M-7B      & -0.8 & -1.1 & 48.2 \\
Q2.5-7B   & -0.6 & -1.6 & 27.6 \\
Q2.5-72B  & -1.6 & -2.3 & 27.6 \\
\hline
Mean      & -1.5 & -2.4 & 33.6 \\
\hline
\end{tabular}
\caption{Full and token-length-matched XTail-minus-Head LP-readout accuracy
changes in percentage points. Retained is the mean percentage of pairs retained
per paradigm.}
\label{tab:length-matched-results}
\end{table}

The aggregate decline persists for every model after matching (Table~\ref{tab:length-matched-results}). Exact-length
matching gives the same conclusion, with a mean decrease of $2.7$ points, but
retains only $20.7\%$ of pairs. Tokenization length therefore does not explain
the Head-to-XTail accuracy difference. Because matching discards many items and
produces model-specific subsets, this analysis is a robustness check rather
than a controlled estimate of the frequency effect.

\subsection{Attractor and Distance Effects}
\label{app:attractor-analysis}

Table~\ref{tab:appendix-attractor-penalties} reports the agreement-attractor analysis. Positive penalties mean that the attractor or distance condition is harder than its no-attractor or local baseline.

Rare lexicalization does not amplify relative-clause subject--verb interference under LP readout: the median penalty is, in fact, highest in Head and lower in Tail and XTail. The same direction appears under Template LP, with smaller absolute penalties. Relative-clause subject--verb attractors remain difficult in absolute terms, but the median attractor penalty decreases from original BLiMP and the generated Head regime to the lower-frequency regimes ($22.3$ points in original BLiMP, $24.3$ in Head, $18.7$ in Tail, and $17.1$ in XTail). Relative-clause attractor accuracy also improves from Head to XTail for all seven base models, with the same direction under Template LP, though with smaller effects.

The pattern is consistent with rare lexicalization removing some high-frequency lexical traps rather than making structural agreement easier. For example, a Head item such as \emph{This lieutenant who had shoved most sergeants \underline{has}/*\underline{have} incorporated some takeovers} places the incorrect auxiliary after a plausible plural local continuation (\emph{sergeants have}). The corresponding XTail lexicalization, \emph{This gamine who had homogenized most zombis \underline{has}/*\underline{have} whopped some galangals}, preserves the same singular head noun, plural attractor, and auxiliary contrast, but makes the local lexical continuation much less familiar. Thus, rare lexicalization does not make models more vulnerable to structural agreement interference. Instead, it appears to reduce accidental lexical or collocational traps present in high-frequency lexicalizations, while leaving overt agreement cues available.

\begin{table*}[t]
\centering
\begin{tabular}{llrrrr}
\hline
\textbf{Domain} & \textbf{Penalty} & \textbf{Orig.} & \textbf{Head} & \textbf{Tail} & \textbf{XTail} \\
\hline
Subject--verb & relative-clause attractor, LP & 22.3 & 24.3 & 18.7 & 17.1 \\
Subject--verb & relative-clause attractor, Temp. LP & 4.0 & 10.3 & 5.9 & 4.0 \\
Subject--verb & relational-noun attractor, LP & 7.9 & 5.3 & 0.1 & 4.9 \\
Determiner--noun & adjective distance, LP & 3.9 & 5.8 & 3.1 & 4.3 \\
\hline
\end{tabular}
\caption{Median attractor or distance penalties in percentage points across seven base models. Positive values indicate that the target condition is harder than the local baseline.}
\label{tab:appendix-attractor-penalties}
\end{table*}

\section{Full Phenomenon and Paradigm-Level Results}
\label{app:phenomenon-tables}

Table~\ref{tab:appendix-phenomenon-results} reports the broad BLiMP phenomenon results, averaging LP-readout scores across the seven base models.

\begin{table*}[t]
\centering
\setlength{\tabcolsep}{4pt}
\begin{tabular}{@{}lrrrrrrrr@{}}
\hline
& \multicolumn{4}{c}{\textbf{LP readout (7 models)}}
    & \multicolumn{4}{c}{\textbf{Yes/No (6 models)}} \\
\textbf{Phenomenon} & \textbf{H} & \textbf{T} & \textbf{XT} &
\textbf{$\Delta$ (SD)} & \textbf{H} & \textbf{T} &
\textbf{XT} & \textbf{$\Delta$ (SD)} \\
\hline
argument struct. & 74.8 & 76.0 & 70.7 & -4.1 (1.8) & 87.3 & 83.1 & 77.3 & -10.0 (2.1) \\
anaphor agr. & 97.6 & 93.4 & 92.7 & -4.9 (0.7) & 97.8 & 91.4 & 92.2 & -5.6 (1.7) \\
island effects & 62.8 & 61.8 & 59.3 & -3.5 (2.0) & 64.4 & 62.4 & 59.0 & -5.3 (2.2) \\
binding & 78.2 & 77.3 & 77.9 & -0.3 (1.3) & 83.8 & 80.4 & 79.4 & -4.5 (2.3) \\
control/raising & 74.3 & 69.5 & 71.1 & -3.3 (2.1) & 73.1 & 70.5 & 69.1 & -4.0 (3.4) \\
NPI licensing & 75.6 & 74.2 & 74.1 & -1.5 (1.2) & 85.5 & 83.6 & 82.5 & -3.0 (3.9) \\
quantifiers & 74.9 & 74.3 & 73.2 & -1.7 (2.4) & 80.2 & 78.9 & 78.1 & -2.1 (2.6) \\
s-selection & 67.1 & 61.9 & 64.3 & -2.8 (1.5) & 75.1 & 73.3 & 73.9 & -1.3 (2.8) \\
irregular forms & 88.6 & 87.1 & 88.1 & -0.5 (1.2) & 96.8 & 96.5 & 95.6 & -1.2 (1.0) \\
ellipsis & 81.0 & 78.3 & 76.1 & -4.9 (4.5) & 87.9 & 88.7 & 86.9 & -1.0 (4.4) \\
det.-noun agr. & 92.5 & 93.4 & 93.2 & +0.7 (0.5) & 92.8 & 94.2 & 93.1 & +0.4 (1.9) \\
filler-gap & 72.9 & 72.6 & 72.2 & -0.7 (0.6) & 69.6 & 71.2 & 70.4 & +0.7 (1.4) \\
subj.-verb agr. & 83.4 & 87.2 & 85.8 & +2.5 (1.2) & 92.0 & 94.0 & 94.4 & +2.4 (1.6) \\
\hline
\end{tabular}
\caption{Broad phenomenon-level accuracy, sorted by mean Yes/No
XTail-minus-Head change. LP averages seven base models; Yes/No averages six
base models, excluding G4-E4B. H, T, and XT are mean accuracies (\%).
$\Delta$ is mean XTail minus Head in percentage points, and parentheses give
the SD across model-specific phenomenon-level changes.}
\label{tab:appendix-phenomenon-results}
\end{table*}

Tables~\ref{tab:appendix-full-paradigm-results-1}--\ref{tab:appendix-full-paradigm-results-4} report the full 67-paradigm breakdown, averaged across the seven base models under LP readout. We sort paradigms by XTail--Head accuracy change, from the largest negative changes to the largest positive changes. The released analysis files include the corresponding full identifiers and model-specific results.

\setlength{\tabcolsep}{3pt}
\begin{table*}[t]
\centering
\begin{tabular}{@{}p{.25\linewidth}rrrrrrrr@{}}
\hline
& \multicolumn{4}{c}{\textbf{LP readout (7 models)}}
    & \multicolumn{4}{c}{\textbf{Yes/No (6 models)}} \\
\textbf{Paradigm} & \textbf{H} & \textbf{T} &
\textbf{XT} & \textbf{$\Delta$ (SD)} & \textbf{H} & \textbf{T} &
\textbf{XT} & \textbf{$\Delta$ (SD)} \\
\hline
Principle A domain 3 & 66.4 & 54.5 & 55.6 & -10.7 (5.1) & 90.5 & 74.9 & 72.6 & -17.9 (3.9) \\
passive 1 & 80.4 & 74.9 & 65.8 & -14.7 (4.0) & 91.4 & 82.9 & 76.2 & -15.1 (2.7) \\
drop argument & 67.3 & 60.0 & 64.6 & -2.7 (3.7) & 87.0 & 74.9 & 72.6 & -14.4 (5.0) \\
passive 2 & 81.6 & 78.4 & 71.4 & -10.2 (3.0) & 93.5 & 84.9 & 80.0 & -13.5 (1.0) \\
expletive there subj. raising & 86.0 & 85.5 & 88.6 & +2.6 (4.7) & 88.5 & 80.1 & 76.1 & -12.4 (12.0) \\
causative & 73.8 & 84.2 & 74.1 & +0.3 (3.4) & 93.1 & 91.1 & 83.9 & -9.2 (4.0) \\
anaphor gender agr. & 97.4 & 90.8 & 90.1 & -7.3 (1.1) & 97.9 & 87.7 & 89.4 & -8.5 (2.2) \\
inchoative & 84.1 & 93.2 & 84.5 & +0.4 (3.1) & 89.6 & 90.6 & 81.5 & -8.1 (6.4) \\
left-branch simple & 76.6 & 75.7 & 69.2 & -7.4 (3.1) & 79.1 & 76.7 & 71.9 & -7.2 (9.5) \\
wh/that with gap & 24.7 & 24.2 & 23.1 & -1.6 (1.9) & 51.6 & 46.8 & 44.8 & -6.8 (2.4) \\
left-branch echo & 47.4 & 49.7 & 45.6 & -1.9 (6.9) & 60.0 & 57.0 & 53.8 & -6.2 (3.4) \\
only NPI scope & 74.9 & 75.1 & 77.1 & +2.2 (1.0) & 68.1 & 65.2 & 62.0 & -6.1 (9.0) \\
wh-island & 65.5 & 63.2 & 58.5 & -7.0 (3.4) & 59.6 & 58.4 & 53.5 & -6.1 (4.0) \\
tough vs. raising 1 & 46.8 & 27.3 & 22.4 & -24.4 (5.6) & 46.2 & 43.1 & 40.3 & -5.9 (10.3) \\
Principle A domain 2 & 78.3 & 75.7 & 78.6 & +0.3 (2.7) & 91.6 & 86.6 & 85.8 & -5.9 (3.7) \\
animate subj. trans. & 70.9 & 61.1 & 63.7 & -7.2 (2.1) & 82.2 & 78.8 & 76.8 & -5.5 (3.3) \\
complex NP island & 52.1 & 49.9 & 48.2 & -4.0 (2.8) & 46.5 & 45.4 & 41.1 & -5.4 (3.4) \\
\hline
\end{tabular}
\caption{Full paradigm-level accuracy results, sorted by mean Yes/No XTail-minus-Head change. LP averages seven base models; Yes/No averages six base models, excluding G4-E4B. $\Delta$ is XTail minus Head, and parentheses give the SD across model-specific changes. Paradigm names are abbreviated for space.}
\label{tab:appendix-full-paradigm-results-1}
\end{table*}

\begin{table*}[t]
\centering
\begin{tabular}{@{}p{.25\linewidth}rrrrrrrr@{}}
\hline
& \multicolumn{4}{c}{\textbf{LP readout (7 models)}}
    & \multicolumn{4}{c}{\textbf{Yes/No (6 models)}} \\
\textbf{Paradigm} & \textbf{H} & \textbf{T} &
\textbf{XT} & \textbf{$\Delta$ (SD)} & \textbf{H} & \textbf{T} &
\textbf{XT} & \textbf{$\Delta$ (SD)} \\
\hline
transitive & 69.9 & 69.4 & 67.3 & -2.6 (2.7) & 83.6 & 83.7 & 78.3 & -5.3 (2.8) \\
sentential subj. island & 41.2 & 37.0 & 34.8 & -6.5 (3.2) & 54.2 & 50.4 & 48.9 & -5.3 (11.7) \\
NPI present 2 & 72.8 & 69.9 & 69.8 & -3.0 (4.0) & 88.9 & 84.5 & 83.6 & -5.3 (3.9) \\
intransitive & 66.5 & 71.8 & 67.1 & +0.6 (1.5) & 72.8 & 73.4 & 68.4 & -4.4 (3.8) \\
coord. struct., complex LB & 70.7 & 70.3 & 68.8 & -1.9 (3.8) & 68.3 & 66.4 & 64.0 & -4.4 (5.4) \\
there quantifiers 2 & 35.9 & 35.9 & 35.4 & -0.5 (4.5) & 74.4 & 72.0 & 70.1 & -4.3 (5.1) \\
coord. struct., obj. extr. & 67.1 & 66.3 & 66.5 & -0.6 (2.3) & 76.6 & 74.1 & 72.4 & -4.2 (4.4) \\
Principle A domain 1 & 97.6 & 97.9 & 97.5 & 0.0 (1.4) & 91.8 & 90.3 & 87.8 & -4.0 (5.8) \\
adjunct island & 81.4 & 82.0 & 82.7 & +1.3 (1.2) & 70.8 & 70.8 & 66.8 & -3.9 (6.2) \\
ellipsis N-bar 1 & 62.8 & 56.8 & 52.7 & -10.1 (8.9) & 91.3 & 90.5 & 87.6 & -3.7 (5.6) \\
wh/that with gap, long & 13.5 & 13.3 & 12.6 & -0.9 (2.4) & 36.0 & 35.3 & 32.6 & -3.4 (2.7) \\
expletive it obj. raising & 84.2 & 73.9 & 79.2 & -5.0 (3.2) & 73.6 & 67.7 & 70.2 & -3.4 (5.9) \\
NPI present 1 & 64.7 & 62.9 & 62.2 & -2.6 (3.4) & 85.6 & 82.7 & 82.5 & -3.2 (3.8) \\
Principle A reconstruction & 35.9 & 36.5 & 36.9 & +1.0 (2.7) & 35.1 & 33.5 & 32.0 & -3.1 (2.7) \\
only NPI licensor & 89.2 & 85.4 & 84.1 & -5.1 (4.4) & 93.9 & 91.6 & 90.9 & -2.9 (5.5) \\
anaphor number agr. & 97.8 & 96.0 & 95.2 & -2.6 (1.0) & 97.7 & 95.1 & 95.0 & -2.7 (1.4) \\
superlative quant. 1 & 88.9 & 85.5 & 83.6 & -5.3 (4.5) & 83.0 & 82.2 & 80.5 & -2.5 (7.1) \\
\hline
\end{tabular}
\caption{Full paradigm-level LP and Yes/No accuracy results, continued.}
\label{tab:appendix-full-paradigm-results-2}
\end{table*}

\begin{table*}[t]
\centering
\begin{tabular}{@{}p{.25\linewidth}rrrrrrrr@{}}
\hline
& \multicolumn{4}{c}{\textbf{LP readout (7 models)}}
    & \multicolumn{4}{c}{\textbf{Yes/No (6 models)}} \\
\textbf{Paradigm} & \textbf{H} & \textbf{T} &
\textbf{XT} & \textbf{$\Delta$ (SD)} & \textbf{H} & \textbf{T} &
\textbf{XT} & \textbf{$\Delta$ (SD)} \\
\hline
Principle A case 2 & 88.5 & 88.0 & 86.6 & -1.9 (3.4) & 95.1 & 93.9 & 92.9 & -2.2 (2.2) \\
matrix question NPI & 65.4 & 64.7 & 63.9 & -1.5 (4.6) & 96.3 & 94.5 & 94.2 & -2.1 (2.0) \\
det.-noun agr. 2 & 96.2 & 94.8 & 92.9 & -3.3 (1.4) & 94.0 & 94.2 & 92.3 & -1.7 (4.0) \\
wh/that no gap & 96.0 & 95.6 & 94.9 & -1.1 (1.1) & 85.3 & 86.0 & 83.8 & -1.6 (1.6) \\
irreg. participle adj. & 91.8 & 87.7 & 90.3 & -1.5 (1.5) & 96.8 & 96.4 & 95.6 & -1.2 (1.0) \\
irreg. participle verbs & 85.4 & 86.5 & 85.9 & +0.4 (1.0) & 96.8 & 96.6 & 95.6 & -1.2 (1.2) \\
superlative quant. 2 & 76.3 & 77.8 & 77.2 & +1.0 (1.4) & 75.6 & 75.0 & 74.5 & -1.2 (7.4) \\
det.-noun adj. 2 & 91.3 & 91.2 & 86.9 & -4.4 (2.9) & 89.5 & 91.6 & 88.6 & -0.9 (5.0) \\
negation NPI licensor & 98.6 & 97.8 & 97.8 & -0.8 (0.6) & 85.9 & 85.3 & 85.1 & -0.8 (8.5) \\
det.-noun agr. 1 & 98.3 & 97.5 & 96.8 & -1.5 (0.6) & 93.9 & 94.9 & 93.3 & -0.7 (1.1) \\
there quantifiers 1 & 98.5 & 98.0 & 96.5 & -2.1 (2.3) & 87.8 & 86.4 & 87.2 & -0.6 (4.6) \\
negation NPI scope & 63.4 & 63.7 & 63.8 & +0.4 (5.8) & 79.8 & 81.3 & 79.5 & -0.4 (4.3) \\
Principle A case 1 & 100.0 & 99.9 & 99.7 & -0.3 (0.3) & 95.4 & 95.6 & 95.1 & -0.3 (2.9) \\
wh/that no gap, long & 95.5 & 95.6 & 96.0 & +0.5 (1.2) & 80.6 & 81.0 & 80.4 & -0.3 (3.3) \\
det.-noun irregular 1 & 93.0 & 94.4 & 95.8 & +2.9 (1.3) & 96.3 & 96.5 & 96.6 & +0.3 (0.7) \\
tough vs. raising 2 & 77.5 & 90.4 & 89.7 & +12.2 (3.9) & 80.3 & 84.0 & 80.8 & +0.5 (5.0) \\
det.-noun irregular 2 & 92.4 & 93.0 & 93.8 & +1.4 (1.1) & 96.6 & 97.0 & 97.1 & +0.5 (0.8) \\
\hline
\end{tabular}
\caption{Full paradigm-level LP and Yes/No accuracy results, continued.}
\label{tab:appendix-full-paradigm-results-3}
\end{table*}

\begin{table*}[t]
\centering
\begin{tabular}{@{}p{.25\linewidth}rrrrrrrr@{}}
\hline
& \multicolumn{4}{c}{\textbf{LP readout (7 models)}}
    & \multicolumn{4}{c}{\textbf{Yes/No (6 models)}} \\
\textbf{Paradigm} & \textbf{H} & \textbf{T} &
\textbf{XT} & \textbf{$\Delta$ (SD)} & \textbf{H} & \textbf{T} &
\textbf{XT} & \textbf{$\Delta$ (SD)} \\
\hline
regular plural SVA 1 & 92.1 & 90.2 & 90.1 & -2.0 (2.3) & 95.8 & 95.2 & 96.5 & +0.7 (0.9) \\
regular plural SVA 2 & 91.4 & 90.8 & 88.2 & -3.1 (3.2) & 95.8 & 97.3 & 96.7 & +0.9 (1.8) \\
det.-noun adj. 1 & 95.8 & 96.5 & 96.0 & +0.2 (0.6) & 88.7 & 91.3 & 90.0 & +1.2 (2.0) \\
expletive there obj. raising & 77.3 & 70.4 & 75.6 & -1.8 (5.0) & 77.0 & 77.5 & 78.3 & +1.4 (6.4) \\
irreg. plural SVA 1 & 88.1 & 92.1 & 89.3 & +1.2 (2.3) & 93.2 & 94.4 & 94.8 & +1.6 (1.1) \\
ellipsis N-bar 2 & 99.2 & 99.7 & 99.5 & +0.3 (0.3) & 84.4 & 87.0 & 86.1 & +1.6 (4.2) \\
det.-noun adj. irr. 1 & 87.5 & 92.2 & 94.1 & +6.5 (1.8) & 91.2 & 93.6 & 93.1 & +1.9 (2.0) \\
Principle A c-command & 81.0 & 88.7 & 90.4 & +9.4 (1.8) & 87.3 & 88.3 & 89.5 & +2.1 (3.1) \\
distractor agr., rel. noun & 82.2 & 88.3 & 85.6 & +3.4 (2.0) & 92.2 & 93.5 & 94.3 & +2.1 (4.0) \\
det.-noun adj. irr. 2 & 85.2 & 87.8 & 89.1 & +3.9 (1.1) & 91.8 & 94.6 & 94.1 & +2.3 (2.3) \\
animate subj. passive & 63.4 & 62.6 & 65.0 & +1.6 (1.7) & 68.0 & 67.8 & 71.0 & +3.0 (4.4) \\
wh obj. gap & 90.6 & 90.0 & 90.4 & -0.3 (1.3) & 78.1 & 83.3 & 81.7 & +3.6 (3.8) \\
irreg. plural SVA 2 & 80.1 & 87.5 & 87.2 & +7.1 (2.2) & 92.8 & 96.5 & 97.0 & +4.2 (1.3) \\
distractor agr., RC & 66.3 & 74.2 & 74.7 & +8.4 (5.0) & 82.3 & 87.0 & 87.3 & +5.0 (4.0) \\
wh subj. gap & 94.1 & 95.1 & 94.4 & +0.3 (1.5) & 80.2 & 85.0 & 86.5 & +6.3 (1.6) \\
wh subj. gap, long & 95.8 & 94.4 & 93.8 & -2.0 (2.3) & 75.5 & 81.2 & 82.8 & +7.3 (6.3) \\
\hline
\end{tabular}
\caption{Full paradigm-level LP and Yes/No accuracy results, continued.}
\label{tab:appendix-full-paradigm-results-4}
\end{table*}
\normalsize
\setlength{\tabcolsep}{6pt}

\section{Human Validation Details}
\label{app:human-validation}

Section~\ref{subsec:human-validation} describes the annotation design and
reports the primary results. This appendix gives the remaining breakdowns.

\paragraph{Non-decision reasons.} Among non-decisions in the declining arm,
unknown words were cited in $0.0\%$, $1.8\%$, $18.2\%$, and $53.8\%$ of
responses from Original through XTail. The remaining non-decisions cited
ambiguity of the contrast or judged both sentences unacceptable.

\paragraph{Agreement structure.} Under two-of-three agreement, 593 of 720 pairs
($82.4\%$) supported the intended acceptable sentence, 91 had a majority
``cannot decide,'' 25 had a three-way split, and 11 had a majority selecting
the intended unacceptable sentence. Manual review did not support reversing any
of the 11 keys, although some of those pairs had weak lexical contrasts or low
naturalness.

\paragraph{Matched model comparison.} We matched annotated items to pair-level
model scores under Yes/No scoring. Across all 540 generated items, mean
accuracy increased from $85.4\%$ to $88.2\%$ after restricting to the 434
human-supported items. G4-E4B is excluded throughout because it did not
reliably use this scoring interface. Because the four conditions contain
independently sampled sentences and human filtering retains different
proportions of items across conditions, the filtered trajectory in
Table~\ref{tab:human-validation} is descriptive rather than a controlled
estimate of the frequency effect.

\paragraph{Instructions shown to annotators.}
The initial task instructions stated:
\begin{quote}
You will see pairs of sentences. One is written correctly and the other has a
mistake. For this first choice, ignore how strange the meaning is and judge only
whether the sentence is put together correctly.
\end{quote}
The accompanying guide further stated:
\begin{quote}
Some meanings and word combinations are intentionally unusual. Judge whether
each sentence is put together correctly, not whether its meaning is realistic.
\end{quote}
Annotators selected sentence A, sentence B, or ``Can't decide.'' Following a
non-decision, the interface asked ``What made this hard to decide?'' and
provided the options ``I don't know some of these words,'' ``Both seem
correct,'' ``Both seem wrong,'' and ``Other,'' with optional free text.

The intended acceptable sentence was then shown alone. The naturalness
instructions stated:
\begin{quote}
Consider meaning and wording together. Could you imagine a fluent English
speaker plausibly producing this sentence to describe something meaningful?
\end{quote}
After a non-decision, annotators were additionally told:
\begin{quote}
Even though you couldn't choose between the pair, please give your best
impression of the sentence below.
\end{quote}
The five response labels were ``Very unnatural,'' ``Somewhat unnatural,''
``Neutral,'' ``Somewhat natural,'' and ``Completely natural.'' Before beginning,
annotators completed three guided examples illustrating common natural wording,
coherent wording containing a rare word, and grammatical but semantically
unnatural wording.

\paragraph{Annotator recruitment and compensation.}
The annotators were recruited through a professional annotation company,
which was responsible for recruiting, managing, and compensating the
annotators, and for their terms of engagement. We did not recruit or pay
the annotators directly. The company provided limited background
information for each annotator, specifically nationality, countries of
residence, and occupation, which we used only to confirm native English
proficiency; we received no other personal information and report these
details only in aggregate. The annotation fee paid to the company was set
at a sufficiently high level to ensure adequate compensation for the
annotators, taking into account the workload and the professional nature
of the annotation task.

\section{Original--Head diagnostics}
\label{app:original-head-gap}

This appendix provides additional diagnostics for the comparison between
original BLiMP and Head \textsc{FreqBLiMP}. This comparison is not the primary
frequency manipulation, since original BLiMP was not generated by the same
lexicalization pipeline as \textsc{FreqBLiMP}. Instead, it indexes broader
effects of regeneration and lexicalization beyond unigram frequency.

\begin{figure*}[t]
  \centering
  \includegraphics[width=.72\textwidth]{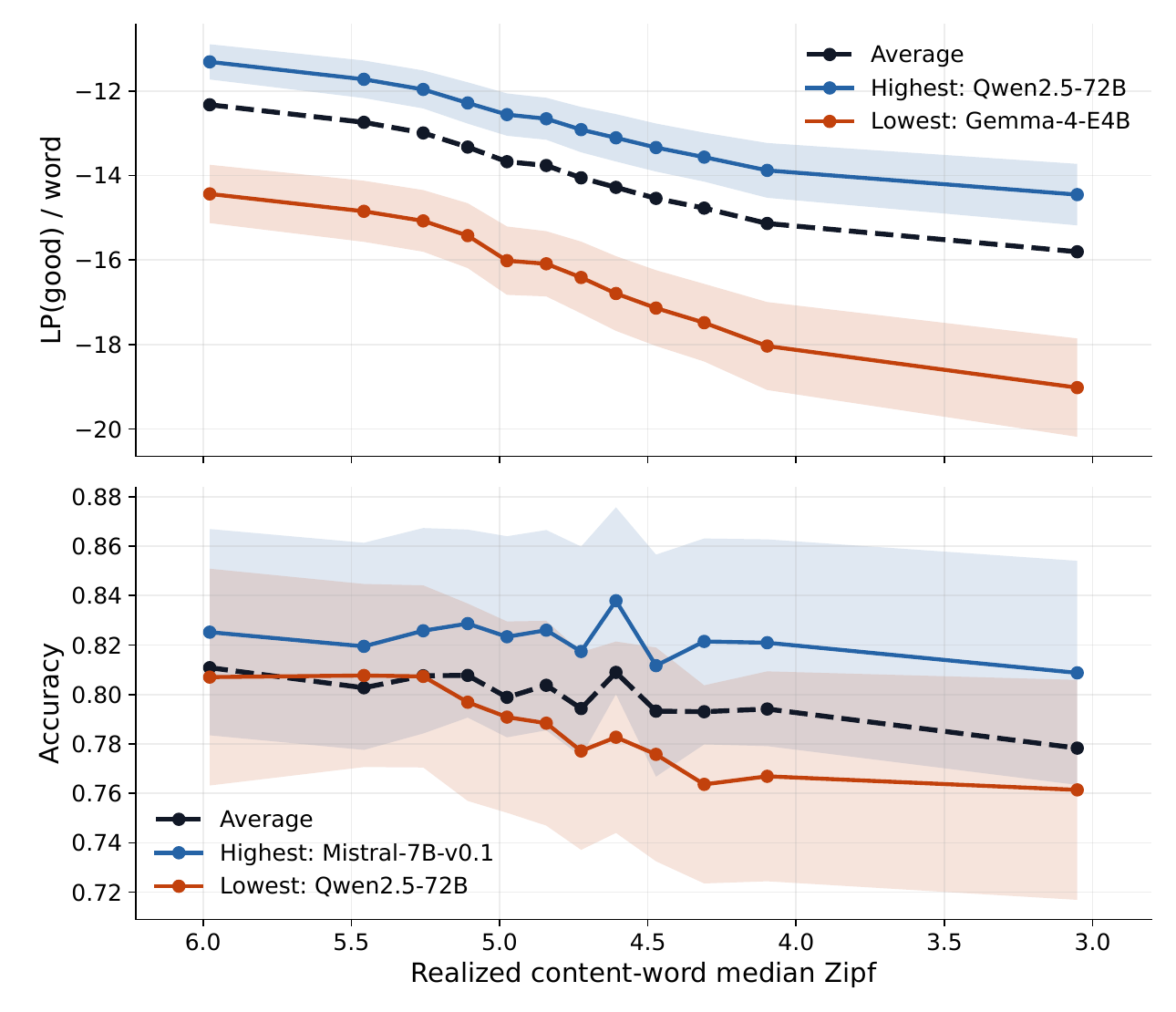}
  \caption{
    Frequency trends within original BLiMP. Top: paradigm-balanced
    word-normalized log probability by realized median Zipf frequency. Bottom:
    paradigm-balanced accuracy by realized median Zipf frequency.
  }
  \label{fig:original-zipf-diagnostics}
\end{figure*}

Figure~\ref{fig:original-zipf-diagnostics} first shows that original BLiMP
itself exhibits the same qualitative likelihood--accuracy dissociation observed
in \textsc{FreqBLiMP}: lower-frequency items receive lower likelihood, while
contrastive accuracy declines only mildly.

Table~\ref{tab:appendix-original-head-margin} gives the direct model-level
comparison, including both accuracy and margin. Here we break the same contrast
down by paradigm.

Table~\ref{tab:largest-original-head-drops} shows that the aggregate
original--Head gap is concentrated rather than diffuse: the median drop across
base models is $1.6$ points, but a small number of paradigms lose $12$ to $24$.
These cluster in filler--gap, island, and tough/raising constructions, where
vocabulary-expanded regeneration can alter the local constructional frame, and
are largely disjoint from the paradigms that decline under increasing lexical
rarity (Table~\ref{tab:localized-paradigms}). Across paradigms, the
correlation between COCA support differences and accuracy differences is modest
(Spearman $\rho=.17$).

For example, in \textsc{tough\_vs\_raising\_1}, a generated grammatical item
such as \textit{All girls are tedious to treasure} competes with an
ungrammatical alternative containing the highly familiar \textit{supposed to}
frame. LP readout can therefore favor local familiarity even when it conflicts
with the intended grammatical contrast.

\begin{table*}[t]
\centering
\begin{tabular}{lrrrrrr}
\hline
\textbf{Model} & \textbf{Orig. acc.} & \textbf{Head acc.} & \textbf{$\Delta$ acc.} & \textbf{Orig. mrg.} & \textbf{Head mrg.} & \textbf{$\Delta$ mrg.} \\
\hline
G4-E4B & 79.9 & 78.3 & -1.6 & 3.51 & 2.83 & -0.68 \\
G4-31B & 78.4 & 78.6 & +0.3 & 3.24 & 2.86 & -0.38 \\
L3.1-8B & 79.6 & 75.7 & -3.9 & 3.34 & 2.34 & -1.01 \\
L3.1-70B & 80.1 & 75.9 & -4.2 & 3.40 & 2.36 & -1.05 \\
M-7B & 81.9 & 78.5 & -3.4 & 3.79 & 2.73 & -1.06 \\
Q2.5-7B & 79.3 & 78.1 & -1.1 & 3.76 & 2.95 & -0.81 \\
Q2.5-72B & 78.5 & 77.3 & -1.2 & 3.63 & 2.81 & -0.82 \\
\hline
\end{tabular}
\caption{Original BLiMP vs. Head \textsc{FreqBLiMP} under LP readout for base models.}
\label{tab:appendix-original-head-margin}
\end{table*}

\begin{table*}[t]
\centering
\setlength{\tabcolsep}{5.0pt}
\begin{tabular}{lrrr}
\hline
\textbf{Paradigm} & \textbf{Original} & \textbf{Head} & \textbf{Gap} \\
\hline
tough vs. raising 1      & 71.0 & 46.8 & -24.3 \\
wh/that with gap, long   & 33.9 & 13.5 & -20.3 \\
coord. struct., obj. extr. & 86.5 & 67.1 & -19.4 \\
wh/that with gap         & 39.5 & 24.7 & -14.8 \\
animate subj. passive    & 75.9 & 63.4 & -12.6 \\
wh-island                & 77.9 & 65.5 & -12.4 \\
\hline
\end{tabular}
\caption{
Paradigms with the largest original--Head LP-readout accuracy drops, averaged
across base models. Values are percentages.
}
\label{tab:largest-original-head-drops}
\end{table*}

\section{Qualitative Examples}
\label{app:qualitative-examples}

Table~\ref{tab:appendix-qualitative-examples} gives examples of lexical
sensitivity, structural robustness, and stimulus ambiguity. Together, they
illustrate why frequency effects should be interpreted as interactions among
rarity, lexical licensing, and local constructional familiarity, rather than
as a pure loss of abstract grammatical knowledge.

\begin{table*}[p]
\centering
\setlength{\tabcolsep}{2.5pt}
\begin{tabular}{@{}
>{\raggedright\arraybackslash}p{.16\textwidth}
>{\raggedright\arraybackslash}p{.13\textwidth}
>{\raggedright\arraybackslash}p{.265\textwidth}
>{\raggedright\arraybackslash}p{.265\textwidth}
>{\raggedright\arraybackslash}p{.13\textwidth}
@{}}
\toprule
\textbf{Role / regime} & \textbf{Paradigm} &
\textbf{Intended acceptable} & \textbf{Intended unacceptable} & \textbf{Interpretation} \\
\midrule
Lexical licensing, Head & causative &
\textit{Paula \underline{closed} some box.} &
$^\ast$\textit{Paula \underline{emerged} some box.} &
Causativization is verb-specific. \\

Orig.--Head gap, Head & wh-island &
\textit{Who might Bill find out \underline{he} stocks?} &
$^\ast$\textit{Who might Bill find out \underline{what} stocks?} &
The island contrast is construction-sensitive. \\

Structural contrast, Head & ellipsis N-bar 1 &
\textit{This controller had adopted three \underline{strict studios} and Teresa had adopted two.} &
$^\ast$\textit{This controller had adopted three studios and Teresa had adopted two \underline{strict}.} &
N-bar ellipsis leaves the adjective implicit. \\

Lexical knowledge, Tail & anaphor gender agr. &
\textit{Every opa isn't slicing \underline{himself}.} &
$^\ast$\textit{Every opa isn't slicing \underline{itself}.} &
\textit{Opa} denotes a grandfather but was unfamiliar to annotators. \\

Lexical licensing, XTail & drop argument &
\textit{Shysters \underline{thieve}.} &
$^\ast$\textit{Shysters \underline{chasten}.} &
Object omission is verb-specific. \\

Structural robustness, XTail & distractor agr., RC &
\textit{A yokel that nationalizes the maskers \underline{chafes}.} &
$^\ast$\textit{A yokel that nationalizes the maskers \underline{chafe}.} &
Singular agreement survives a plural attractor. \\

Structural robustness, XTail & principle A c-command &
\textit{The bambinos who are slating the nightie weren't denuding \underline{themselves}.} &
$^\ast$\textit{The bambinos who are slating the nightie weren't denuding \underline{itself}.} &
Overt reflexive number remains recoverable. \\

Human uncertainty, XTail & inchoative &
\textit{Those zoologists \underline{congeal}.} &
$^\ast$\textit{Those zoologists \underline{assail}.} &
The verb-frame key is valid, but the good sentence is implausible. \\

Weak selection, XTail & animate subj. trans. &
\textit{\underline{This shamash} exhorts the chartists.} &
$^\ast$\textit{\underline{Every epigraph} exhorts the chartists.} &
Rare word senses weaken the animacy contrast. \\
\bottomrule
\end{tabular}
\renewcommand{\arraystretch}{1}
\caption{
Examples illustrating lexical sensitivity, structural robustness, and
stimulus ambiguity. Underlining marks the local contrast driving the minimal
pair. Intended labels are used because the audit cases need not receive
categorical human judgments.
}
\label{tab:appendix-qualitative-examples}
\end{table*}

\end{document}